\documentclass[10pt,twocolumn,letterpaper]{article}

\usepackage{iccv}              

\definecolor{iccvblue}{rgb}{0.21,0.49,0.74}
\usepackage[pagebackref,breaklinks,colorlinks,allcolors=iccvblue]{hyperref}
\usepackage{amsmath}
\usepackage{multirow}
\usepackage{pifont}
\usepackage{graphicx}
\usepackage{tabularx}
\usepackage{comment}
\usepackage{cuted}
\usepackage[table,xcdraw]{xcolor} 

\usepackage{booktabs}
\usepackage{pifont} 

\newcommand{\cmark}{\ding{51}}  
\newcommand{\xmark}{\ding{55}}  

\definecolor{mygray}{RGB}{230,230,230}

\usepackage{etoc}

\def\paperID{6707} 
\def\confName{ICCV}
\def\confYear{2025}

\title{How Far are AI-generated Videos from Simulating the 3D Visual World:\\A Learned 3D Evaluation Approach}

\author{
Chirui Chang$^{1}$ \quad 
Jiahui Liu$^{1}$ \quad 
Zhengzhe Liu$^{3}$ \quad 
Xiaoyang Lyu$^{1}$ \quad 
Yi-Hua Huang$^{1}$ \\
Xin Tao$^{2}$ \quad
Pengfei Wan$^{2}$ \quad
Di Zhang$^{2}$ \quad 
Xiaojuan Qi$^{1}$\thanks{Corresponding author.} \\
$^{1}$The University of Hong Kong \quad
$^{2}$Kling Team, Kuaishou Technology \quad
$^{3}$Lingnan University
}
\begin{document}
\maketitle

\begin{strip}
     \centering
     \includegraphics[width=0.95\linewidth]{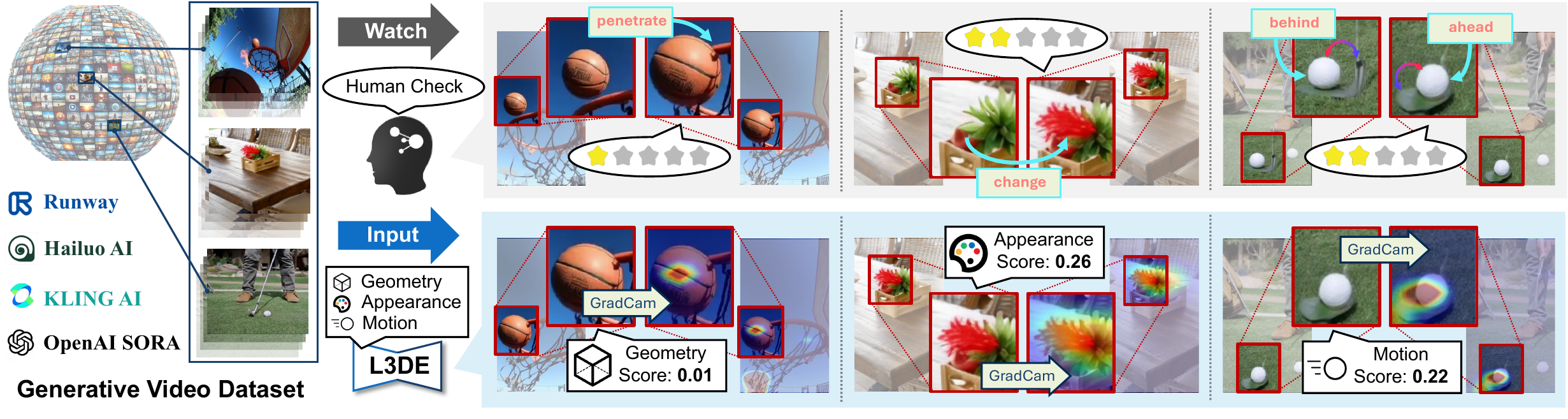}
     \captionof{figure}{L3DE evaluates videos from any generative model based on 3D visual coherence, assessing appearance, motion, and geometry. Its scores align closely with human perception and can localize regions of 3D simulation failures, similar to human intuition. Examples highlight key failure cases: (1) incorrect occlusion between the basketball and hoop, disrupting geometric consistency, (2) abrupt texture transition in plant leaves, and (3) unnatural relative motion between the golf ball and the golf club, violating real-world motion dynamics.
     }
     \label{fig:teaser}
\end{strip}

\begin{abstract}
Recent advancements in video diffusion models enable the generation of photorealistic videos with impressive 3D consistency and temporal coherence. However, the extent to which these AI-generated videos simulate the 3D visual world remains underexplored. 
In this paper, we introduce Learned 3D Evaluation (L3DE), an objective, quantifiable, and interpretable method for assessing AI-generated videos’ ability to simulate the real world  in terms of 3D visual qualities and consistencies, without requiring manually labeled defects or quality annotations.
Instead of relying on 3D reconstruction, which is prone to failure with in-the-wild videos, L3DE employs a 3D convolutional network, trained on monocular 3D cues of motion, depth, and appearance, to distinguish real from synthetic videos.
Confidence scores from L3DE quantify the gap between real and synthetic videos in terms of 3D visual coherence, while a gradient-based visualization pinpoints unrealistic regions, improving interpretability. 
We validate L3DE through extensive experiments, demonstrating strong alignment with 3D reconstruction quality and human judgments. Our evaluations on leading generative models (e.g., Kling, Sora, and MiniMax) reveal persistent simulation gaps and subtle inconsistencies.
Beyond generative video assessment, L3DE extends to broader applications: benchmarking video generation models, serving as a deepfake detector, and enhancing video synthesis by inpainting flagged inconsistencies. 
\end{abstract}    
\section{Introduction}
Video diffusion models, such as Sora \cite{videoworldsimulators2024}, have recently shown remarkable capabilities in visual simulation, producing photorealistic videos with 3D consistency and temporal coherence that can even deceive human observers. This progress raises a fundamental question: how well do AI-generated videos simulate the 3D visual world? While existing evaluations heavily rely on subjective user studies, a quantifiable and interpretable approach remains missing for assessing 3D visual coherence of generative videos.
3D scene reconstruction~\cite{mildenhall2021nerf, kerbl20233d, huang2024sc, wu20244d, yang2024deformable} is a natural way to assess whether generative videos preserve 3D visual coherence.
The intuition is that if a video enables high-quality 3D reconstruction, it should maintain 3D-consistent appearance, structure, and motion across frames. 
However, even state-of-the-art reconstruction methods struggle with in-the-wild videos due to challenges such as unreliable pose estimation~\cite{yang2024deformable, fan2024instantsplat, wang2024dust3r} and the absence of multi-view cues~\cite{voleti2024sv3d,fan2024instantsplat,lei2024mosca,zhang2021ners}, making large-scale evaluation based on reconstruction impractical.
To overcome these limitations, inspired by~\cite{ferwerda2003three}, we turn to monocular 3D cues, such as depth and optical flow, which naturally emerge from videos and serve as strong proxies for 3D structure and motion. Thus, we explore leveraging monocular cues from foundation models \cite{piccinelli2024unidepth, teed2020raft} as an alternative for assessing 3D realism. Specifically, we use RAFT \cite{teed2020raft} for optical flow estimation and UniDepth \cite{piccinelli2024unidepth} for depth prediction, while utilizing DINOv2~\cite{oquab2023dinov2} to capture high-level appearance features.

We collect real and synthetic videos from Pexels~\cite{Pexels} and Stable Video Diffusion (SVD)~\cite{blattmann2023stable}, respectively. Pexels provides diverse real-world videos, while SVD is one of the most accessible video generator. We align their visual content by using real video frames as prompts to generate paired synthetic videos. This minimizes content disparities, isolating differences in 3D consistency and helping analyze how generative videos deviate from real ones.

Equipped with 3D proxies and data, the next challenge is measuring the gap between generative and real-world videos. To tackle this, we develop \textbf{Learned 3D Evaluation (L3DE)}, a data-driven learning-based tool that uses monocular 3D cues to evaluate generative videos and identify 3D visual simulation failures.
L3DE captures intrinsic differences between real and synthetic videos by training a 3D convolutional network with contrastive learning using 3D proxies as inputs. The confidence scores quantify the gap between synthetic and real videos regarding these 3D proxies.
Additionally, L3DE enhances interpretability by highlighting key failure regions via a gradient-based method \cite{selvaraju2016grad} (see Table~\ref{tab:huamn_ps}).
Finally, by integrating depth, motion, and appearance proxies through a feature fusion module, L3DE provides a stable and comprehensive evaluation of 3D visual coherence in generative videos.

To validate L3DE’s effectiveness, we conduct 3D scene reconstruction experiments and user studies. 
Our results in Sec.~\ref{sec:reconstruction_align} show that L3DE scores highly correlate with reconstruction quality, with flagged areas aligning with regions of high 3D inconsistency, as confirmed by reconstruction errors.
Human studies in Sec.~\ref{sec:human_align} further reveal that L3DE scores align closely with human perceptual judgments, with flagged areas consistently rated high by annotators. These results demonstrate L3DE’s effectiveness in assessing and analyzing 3D visual coherence in generative videos.
We conduct experiments applying L3DE to videos from leading generative models, including Sora \cite{videoworldsimulators2024}, MiniMax \cite{minimax2024hailuo}, Kling \cite{kuaishou2024kling}, and others to benchmark their 3D visual simulation capabilities and analyze their strengths and limitations.
With L3DE validated through 3D reconstruction and human evaluation, these results provide insights into how well different models capture 3D realism.
As shown in Table \ref{tab:evaluation}, models like Sora and Kling achieve higher L3DE scores, particularly in appearance simulation, while all models show room for improvement in motion and geometry consistency.
Most generative videos still exhibit noticeable gaps from real ones in 3D visual coherence, as reflected in their lower L3DE scores.
Beyond evaluating 3D visual coherence in AI-generated videos, L3DE can serve as a deepfake detector by applying a confidence score threshold. Despite not being trained on videos from specific sources, L3DE effectively identifies fake videos from Kling and others (see Table 1 in the appendix) with over 0.7 accuracy.
Additionally, L3DE’s localized failure regions can help improve video synthesis. By inpainting flagged areas (see appendix), we can enhance the 3D visual coherence of generative videos.

Our contributions can be summarized as follows:

\begin{itemize}[noitemsep,nolistsep]

\item We take the first step in systematically investigating the 3D visual coherence of AI-generated videos across appearance, motion, and geometry—key factors in representing a dynamic 3D world.
To facilitate quantitative analysis, we extract monocular clues from foundation models to disentangle these aspects.

\item We introduce Learned 3D Evaluation (\textbf{L3DE}) that quantifies the 3D visual coherence of a video using confidence scores from models trained on pairing data with contrastive loss. L3DE also highlights spatial and temporal regions as evidence for its assessment. Moreover, we integrate these three aspects to deliver a more robust assessment tool.

\item Through controlled user studies and 3D reconstruction experiments on diverse generative videos, we show that L3DE’s quantification scores and localized regions align well with user intent and reconstruction quality.

\item L3DE can be used for broader applications. Our experiments and studies provide valuable insights and findings about the capabilities of current video generation models.

\end{itemize}
\section{Related Work}

\vspace{0.05in}\noindent\textbf{Diffusion models for video generation.}
The success of diffusion models~\cite{ho2020denoising,sohl2015deep} in image synthesis~\cite{dhariwal2021diffusion,rombach2022high,gu2022vector,nichol2021glide,saharia2022photorealistic,podell2023sdxl,liu2024object} has driven advancements in video generation~\cite{blattmann2023align,ge2023preserve,guo2023animatediff,ho2022imagen,ho2022video,khachatryan2023text2video,luo2023videofusion,zhang2023show,xing2023dynamicrafter,zhou2022magicvideo}.
Stable Video Diffusion~\cite{blattmann2023stable} leverages large-scale training for high-quality video synthesis. Sora~\cite{videoworldsimulators2024} demonstrate the ability to simulate humans, animals, and environments, highlighting video generation as a potential path towards world simulation.
Our work aim to help the community gain more understanding about generative videos, especially their gap from real-world videos in terms of 3D viusal simulation capabilities.

\vspace{0.05in}\noindent\textbf{AI-generated video evaluation.}
Existing metrics for evaluating AI-generated videos include Inception Score (IS)~\cite{salimans2016improved}, Fréchet Video Distance (FVD)~\cite{unterthiner2019fvd}, Perceptual Input Conformity(PIC)~\cite{xing2023dynamicrafter} and CLIPSIM~\cite{radford2021learning}, among others. 
Recent benchmarks, such as VBench~\cite{huang2023vbench} and EvalCrafter~\cite{liu2024evalcrafter}, establish standardized protocols by integrating automated metrics for comprehensive model comparisons.
In contrast, our approach identifies differences between real and generative videos using a data-driven yet simple method, complemented by low-level statistical analysis to assess their 3D visual simulation capabilities.

\vspace{0.05in}\noindent\textbf{Video feature extraction.}
Extracting appearance, motion, and geometry information is crucial for evaluating video realism.
DINOv2~\cite{oquab2023dinov2} shows strong image appearance representation, while optical flow estimation methods~\cite{dosovitskiy2015flownet,ilg2017flownet,teed2020raft} provide robust motion features.
Monocular depth cues encode rich geometric information, with recent methods like UniDepth~\cite{piccinelli2024unidepth} achieving precise metric depth estimation with excellent video consistency.
We leverage these techniques to extract relevant features for our analysis.

\vspace{0.05in}\noindent\textbf{3D scene reconstruction.}
Recent advancements in 3D reconstruction, such as NeRF-based~\cite{mildenhall2021nerf,pumarola2021d,wang2021neus,yu2022monosdf,barron2022mip,park2021hypernerf,li2022neural,lyu2024total} and 3D-GS-based~\cite{kerbl20233d,huang2024sc,wu20244d,yang2024deformable} methods, have improved static and dynamic scene modeling.
Despite the robustness of novel view synthesis (NVS) methods for in-the-wild scenes, unreliable camera pose estimation in such videos limits the feasibility of 3D scene reconstruction as a robust large-scale evaluation tool for assessing the 3D visual simulation capabilities of AI-generated videos.
\begin{table*}[ht]
\centering
\renewcommand\arraystretch{1.1}
\resizebox{0.99\linewidth}{!}{%
\begin{tabular}{lcccccc}
\toprule
 \textbf{Source} & \textbf{Synthetic/Real} & \textbf{Number of Videos} & \textbf{Clip Length} & \textbf{Resolution} & \textbf{Frame Rate} & \textbf{Prompt Type}\\
\midrule
\multicolumn{7}{c}{\textbf{Paired Real/Synthetic Video Set}} \\
\midrule
Pexels~\cite{Pexels} & Real & 80,000 & 4s & Variable & Variable & --\\
Stable Video Diffusion~\cite{blattmann2023stable} & Synthetic & 80,000 & 4s & 1024*576 & 7 FPS & I2V\\
\midrule
\multicolumn{7}{c}{\textbf{3D Reconstruction Verification Set}} \\
\midrule
Kling 1.5~\cite{kuaishou2024kling} & Synthetic & 3,000 & 5s & Variable & 30 FPS & I2V \& T2V\\
\midrule
\multicolumn{7}{c}{\textbf{3D Visual Simulation Benchmark}} \\
\midrule
Pexels~\cite{Pexels}& Real & 14,000 & 4s & Variable & Variable & --\\
Runway-Gen3~\cite{runway2024gen3} & Synthetic & 539 & 5s & 1280*768 & 24 FPS & I2V \& T2V\\
MiniMax~\cite{minimax2024hailuo} & Synthetic & 539 & 5s & 1280*720 & 25 FPS & I2V \& T2V\\
Vidu~\cite{shengshu2024vidu} & Synthetic & 539 & 3s & Variable & 24 FPS & I2V \& T2V\\
Luma Dream Machine 1.6~\cite{luma2024dm} & Synthetic & 539 & Variable & Variable & 24 FPS & I2V \& T2V\\
Kling 1.5~\cite{kuaishou2024kling} & Synthetic & 539 & 5s & Variable & 30 FPS & I2V \& T2V\\
CogVideoX-5B~\cite{yang2024cogvideox} & Synthetic & 539 & 6s & 720*480 & 8 FPS & I2V \& T2V\\
Sora~\cite{videoworldsimulators2024} & Synthetic & 539 & 5s & Variable & 30 FPS & I2V \& T2V\\
Kling 2.1~\cite{kuaishou2024kling} & Synthetic & 539 & 5s & Variable & 30 FPS & I2V \& T2V\\
\bottomrule
\end{tabular}
}
\caption{Overview of our dataset, which consists of (1) Paired Real/Synthetic Video Set, designed to study the gap between real-world and AI-generated videos; (2) the 3D Reconstruction Verification Set, curated for validating L3DE through 3D reconstruction; and (3) the 3D Visual Simulation Benchmark, which includes videos from multiple generative models to evaluate their 3D visual simulation capabilities.}
\label{table:dataset_new}
\vspace{-8pt}
\end{table*}
\section{Data Curation}
\label{sec: data}
To gain a deeper understanding of the 3D visual simulation capabilities of AI-generated videos, we design a data curation process and compile a dataset that includes both real-world and AI-generated videos, as detailed in Table \ref{table:dataset_new}. Our model training, method validation, and subsequent analysis are all conducted using different subsets in this dataset.

\vspace{0.05in}\noindent\textbf{In-the-wild real-world videos.}
We begin by collecting approximately 100,000 real-world, in-the-wild videos from Pexels \cite{Pexels}. These videos encompass a wide range of content, including animals, people, natural scenes, urban landscapes, indoor environments, and more.
For raw video processing, we follow the method introduced in \cite{blattmann2023stable}. 
More details on the data processing can be found in the appendix.

\vspace{0.05in}\noindent\textbf{Paired generative videos.}
We employ the open-source generative model Stable Video Diffusion (SVD)~\cite{blattmann2023stable} to generate synthetic videos. To ensure the focus is on the 3D visual coherence, rather than potential biases in the generated content or color distribution, we condition SVD model using the first frames from real video clips. This enables SVD to generate paired synthetic samples that preserve the same semantic content and color distribution as their real video counterparts.
Thus we create a paired generative video dataset, where the video clips share similar visual content to the real videos, minimizing the risk of model bias.

{\vspace{0.05in}\noindent\textbf{3D reconstruction verification set.}  
To evaluate L3DE's effectiveness, we curate a verification set using videos generated by the commercial model Kling \cite{kuaishou2024kling}, as SVD-generated videos are typically of low quality, hindering 3D reconstruction and rendering.  
Our verification set consists of two parts:  
(1)\textit{ Generated Videos for In-the-wild Scenes.} Given the low success rate of pose estimation \cite{yang2024deformable, fan2024instantsplat, wang2024dust3r} on AI-generated videos, we generate diverse samples conditioned on keyframes from unseen real videos. We then screen the large pool of generated videos and retain 30 that successfully undergo 3D reconstruction.  
(2)\textit{ Twin Videos for Public Scene Datasets.} To analyze the correlation between 3D consistency and L3DE score, we iteratively generate twin videos for 15 scenes from public static datasets (i.e., Mip-NeRF360 \cite{barron2022mip}, Tanks-and-Temples \cite{Knapitsch2017}) and dynamic datasets (i.e., Hyper-NeRF \cite{park2021hypernerf}, Neural 3D Video Synthesis Dataset \cite{li2022neural}), ensuring that each scene yields at least one video that successfully undergoes COLMAP and reconstruction. Each twin video pair is generated using one real frame as the start frame and another with sufficient overlap as the end frame, maintaining close alignment with the real 3D content.  
Videos from (1) and (2) form the 3D reconstruction verification set, totaling 3000 videos. For validation experiments, we use only videos that successfully undergo pose estimation, while the entire set is used in supplementary fake video detection experiments.}

\vspace{0.05in}\noindent\textbf{3D visual simulation benchmark.}
 We conduct studies using L3DE on generated videos from recent commercial generative models, augmented with data from \cite{zeng2024dawn}, to assess their ability to simulate the 3D visual world. The dataset includes videos from models such as Sora \cite{videoworldsimulators2024}, Kling \cite{kuaishou2024kling}, Runway-Gen3 \cite{runway2024gen3}, Luma \cite{luma2024dm}, MiniMax \cite{minimax2024hailuo}, Vidu \cite{shengshu2024vidu}, and CogVideoX \cite{yang2024cogvideox}.
To ensure relevance, we exclude videos with non-realistic content, such as animations. 
Since all videos are generated with the same set of image or text prompts, this dataset enables a direct and fair comparison of 3D visual simulation capabilities across different models by eliminating prompt-induced variability.
Furthermore, we provide 14,000 unseen real video samples as references to establish an empirical upper bound for L3DE scores.
 
\section{Learned 3D Evaluation}

Below, we first discuss proxies for representing the 3D visual world, followed by a detailed explanation of the newly proposed Learned 3D Evaluation (L3DE) for assessing the 3D visual simulation capabilities of AI-generated videos.

\subsection{Proxies for Representing 3D Visual World}
\label{sec:representation}

Reconstructing and rendering in-the-wild videos to assess 3D world simulation is challenging, primarily due to issues such as unreliable camera pose estimation \cite{schoenberger2016sfm,schoenberger2016mvs,yang2024deformable}. 
Beyond reconstructing a scene in 3D space, the realism of the 3D visual world is shaped by multiple perceptual factors. Inspired by~\cite{ferwerda2003three, sarkar2024shadows}, we identify three key aspects: 
\textbf{1) Appearance:} Visual attributes of video frames, including color, texture, and lighting; \textbf{2) Motion:} Temporal dynamics and changes within the video; and \textbf{3) Geometry:} The spatial structure and shape of objects in the frames. These cues reflect the consistency of a video’s 3D structure and can be reliably estimated from videos using foundation models, which we leverage as proxies for the 3D visual world. We extract these cues using the following foundation models:

\begin{itemize}
\item[$\bullet$] \textbf{Appearance representation}: Instead of simply using the original RGB information, we extract per-frame visual feature with DINOv2~\cite{oquab2023dinov2} as the appearance representation. 
Its features are capable of cross-image dense and sparse matching \cite{oquab2023dinov2,edstedt2024roma}, which enhances the potential to capture cross-frame appearance consistency. 
\item[$\bullet$] \textbf{Motion representation}: We leverage optical flow, which is well-studied to represent motion, to examine the motion pattern differences between synthetic and real videos.
To be more specific, we employ RAFT~\cite{teed2020raft}, a state-of-the-art optical flow estimation model, to extract optical flow between the adjacent frames. 

\item[$\bullet$] \textbf{Geometry representation}: 
To investigate the geometric properties of generative videos, we leverage the per-frame depth as the geometry representation. 
Depth conveys many 2.5D geometric cues, such as occlusion, spatial relationships, scales, and so on. 
In detail, considering the cross-frame scale consistency, we adopt metric depth from UniDepth~\cite{piccinelli2024unidepth} as it has a uniform scale and provides better consistency across frames, which aids in perceiving changes in the geometric structure of the video.
\end{itemize}

\subsection{Design of L3DE}
\label{sec:l3de}

With the prepared data and extracted 3D visual proxies, we develop L3DE. The model first trains a classifier on the paired real/synthetic video dataset in Table \ref{table:dataset_new}, enabling it to learn to distinguish them based on the three proxies. This is achieved with a contrastive learning objective, which enhances the discriminative power of the learned features. Additionally, we integrate Grad-CAM \cite{selvaraju2016grad} to enable L3DE to identify simulation traits. Finally, we design a fusion module that combines all three 3D proxies to produce a more comprehensive evaluation score for video assessment.

\vspace{0.05in}\noindent\textbf{Classifier construction and training.} 

Based on the 3D proxies outlined in Sec. \ref{sec:representation}, we design a 3D convolutional network with multiple layers interleaved with ReLU activation functions. It predicts the confidence score evaluating whether a sample belongs to real or synthetic videos. Further details are provided in the appendix. The penultimate layer features are used to construct the contrastive loss. For any input generative video feature \( \mathbf{f}_{\text{gen}} \), the loss encourages pushing apart its closest real video feature, thereby making real video feature more distinguishable. 
It is computed as:
\begin{equation}
\mathcal{L}_{\text{contrastive}} = \sum_{i} \exp \left( - \left\| \mathbf{f}_{\text{gen}}^{(i)} - \mathbf{f}_{\text{real}}^{(j(i))} \right\|_2^2 \right),
\label{eq:loss_feat_neg}
\end{equation}
where \( \mathbf{f}_{\text{real}}^{(j(i))} \) is the closest real video feature to \( \mathbf{f}_{\text{gen}}^{(i)} \) in Euclidean distance. The total loss function combines the classification loss \( \mathcal{L}_{\text{cls}} \) and contrastive loss \( \mathcal{L}_{\text{contrastive}} \) as follows:
\begin{equation}
\mathcal{L} = \mathcal{L}_{\text{cls}} + \lambda \mathcal{L}_{\text{contrastive}}.
\label{eq:total_loss}
\end{equation}
As the network learns to distinguish between real and synthetic videos, its confidence scores serve as a quantitative metric for assessing how closely an input video resembles real-world videos in 3D visual coherence. To interpret its predictions and understand the underlying evidence, we apply Grad-CAM \cite{selvaraju2016grad}, which generates a class-discriminative localization map by backpropagating gradients to the last convolutional layer. This map highlights the video regions that mostly influence the model's decision (see Fig.~\ref{fig: validation2}).

\vspace{0.05in}\noindent\textbf{Feature fusion for comprehensive scores. } Since video content inherently combines appearance, motion, and geometry, we design a feature fusion module for a more robust and comprehensive evaluation. Within the network, we concatenate features from these three aspects:
\begin{equation}
\mathbf{f}_{\text{fused}} = \text{Concat}\left( \mathbf{f}_{\text{app}},\ \mathbf{f}_{\text{mot}},\ \mathbf{f}_{\text{geo}} \right),
\label{eq:fused_feature}
\end{equation}
where \( \mathbf{f}_{\text{app}} \), \( \mathbf{f}_{\text{mot}} \), and \( \mathbf{f}_{\text{geo}} \) represent the features for appearance, motion, and geometry. The fused representation in our Fusion variant of L3DE produces an overall score, jointly accounting for all three aspects. This holistic evaluation provides a more comprehensive measure of 3D visual coherence, complementing single-aspect assessments.
\definecolor{mycolor}{rgb}{0.8, 0.8, 0.8}

\section{Validation of L3DE} 
\label{sec: validation}
To validate L3DE’s reliability in evaluating 3D visual coherence, we employ two complementary strategies: 3D reconstruction and human perceptual judgment. 
3D reconstruction objectively assesses how well AI-generated videos preserve spatial structure and motion realism. However, pose estimation often fails on in-the-wild videos, meaning that only a subset of videos-- those where camera parameters can be reliably estimated-- 
 can be reconstructed for validation. Within this subset, we use reconstruction to precisely verify L3DE’s predicted scores and detected regions.
Beyond this subset, human perception provides a more flexible and perceptually grounded evaluation of 3D visual coherence, as it is not constrained by camera estimation failures. This allows us to confirm that L3DE remains effective across a wider range of generative videos.

\subsection{Validation using 3D Reconstruction}
\label{sec:reconstruction_align}
We conduct 3D reconstruction experiments in two controlled settings to assess the correlation between L3DE scores and 3D rendering quality. Additionally, we examine whether L3DE's detected inconsistencies align with reconstruction errors by comparing its localized regions to rendering-based discrepancy maps. These experiments utilize the 3D reconstruction verification dataset (Table \ref{table:dataset_new}).

\begin{table}[t]
    \renewcommand\arraystretch{1.1}
    \resizebox{\linewidth}{!}{
    \begin{tabular}{lcccc}
        \toprule
         \bf Correlation with L3DE & \bf Fusion & \bf Appearance & \bf Motion & \bf Geometry \\ 
        \midrule
        \bf 3D Reconstruction Quality & 0.7566 & 0.7181 & 0.6669 & 0.3142 \\
        \bf Human Ratings & 0.6460 & 0.5643 & 0.4617 & 0.3479 \\
        \bottomrule
    \end{tabular}
        }
    
    \caption{Spearman correlation between L3DE scores and different reference evaluations. The first row shows correlation with 3D reconstruction quality, while the second shows correlation with human ratings on the same verification dataset.}
    \label{tab:cor_render}

\end{table}

\begin{figure}[t]
    \centering
    \includegraphics[width=\columnwidth]{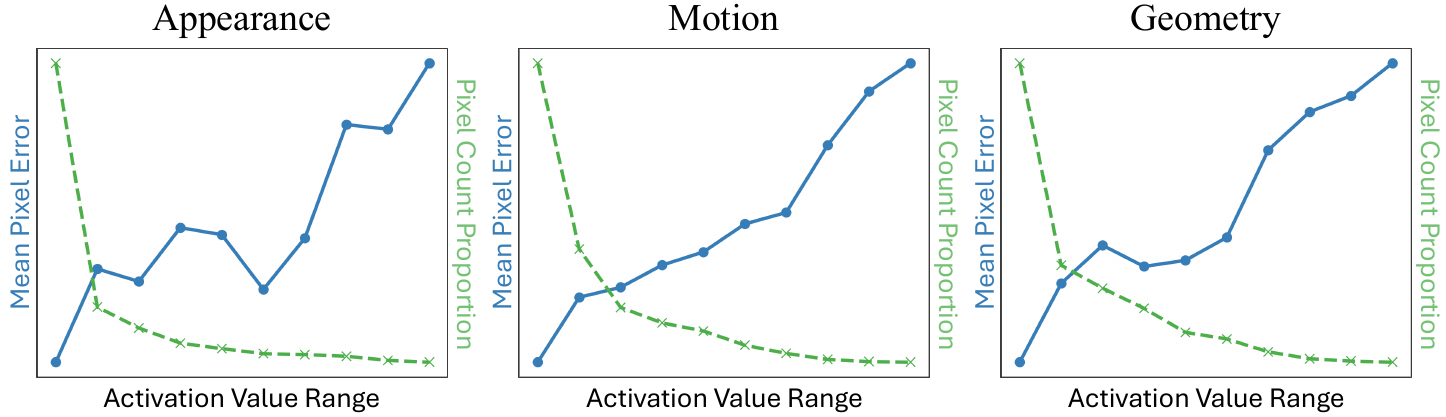}
    \caption{Illustration of the statistics of activation value, pixel value error and the distribution of pixel number for each proxy.}
    \vspace{-8pt}
    \label{fig:cam_val}
\end{figure}

\begin{figure*}
  \centering
        \includegraphics[width=0.98\linewidth]{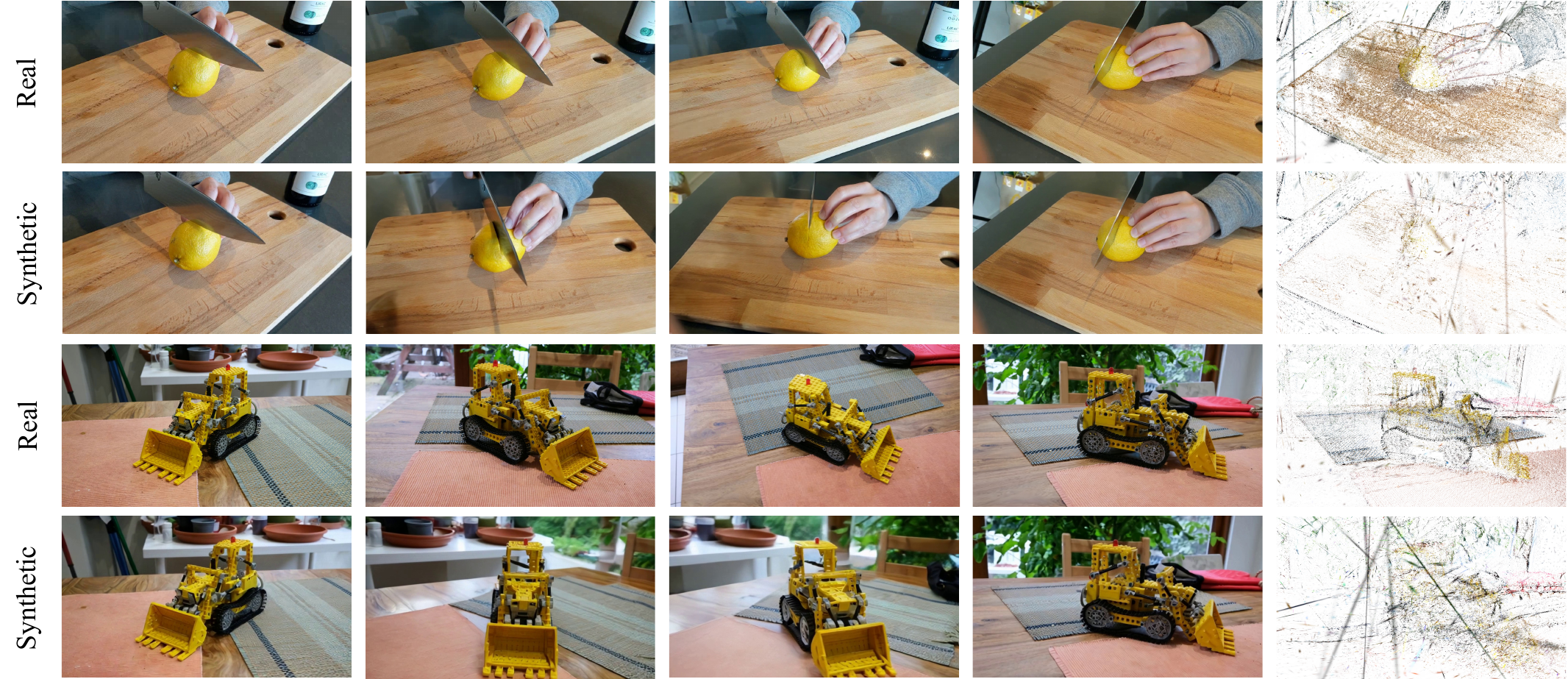}
    \caption{Frames and reconstruction results of twin videos. Even though synthetic videos appear plausible, they do not achieve the same level of 3D scene reconstruction accuracy as real videos (see the Shrunken Gaussians in the rightmost column). This discrepancy underscores a key limitation: current generative videos are not yet adept at faithfully simulating the world in terms of 3D visual coherence.}
    \label{fig: validation1} 
    \vspace{-6pt}
\end{figure*}

\begin{figure*}
  \centering
        \includegraphics[width=0.98\linewidth]{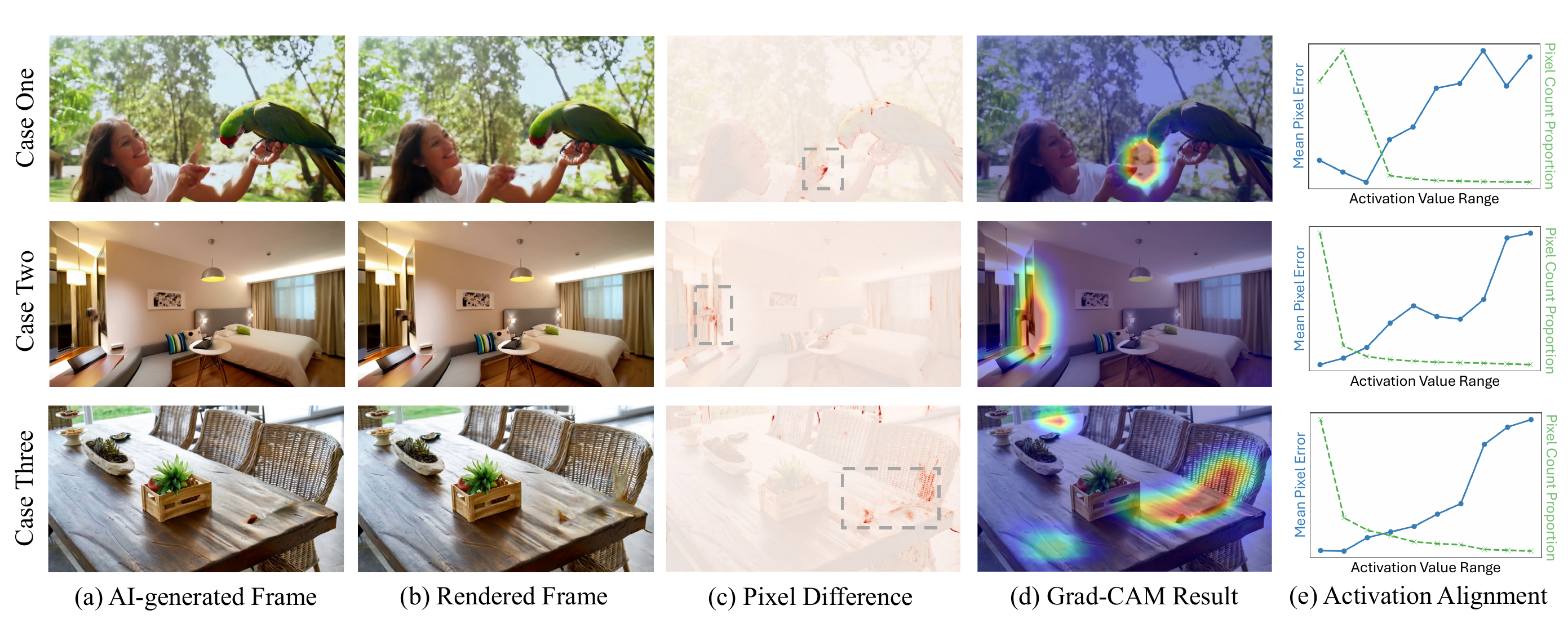}
    \caption{Illustration of 3D inconsistencies identified by L3DE. From left to right: (a) AI-generated video frame; (b) rendered frame with 3D reconstruction with pose aligned with the original view; (c) pixel-level difference between (a) and (b); (d) Grad-CAM result from the L3DE network, which closely aligns with (c); (e) Blue solid line: large (normalized) activation value in (d) is highly aligned with large mean pixel value error in (c). Green dashed line: areas with high (normalized) activation values cover only a small portion of the entire frame. L3DE identifies key artifacts in the cases: (1) unnatural hand motion in the first case, reflected in a low motion score of 0.4642; (2) abrupt geometric deformation of the marked object in the second case, with a geometry score of 0.637; and (3) sudden texture changes in the chair and table in the third case, resulting in an appearance score of 0.2578.
    }
    \label{fig: validation2} 
    \vspace{-6pt}
\end{figure*}

\vspace{0.05in}\noindent\textbf{L3DE score v.s. 3D rendering quality.}
We evaluate the correlation between L3DE scores and 3D reconstruction quality by optimizing a 3D representation, such as 3D-GS \cite{kerbl20233d}, across all video frames to reconstruct each scene.
To ensure a more adaptive evaluation, we use the ‘Twin Videos for Public Scene Datasets’ from the 3D reconstruction verification set, which provides real and synthetic videos of the same content for fair comparisons.
For static scenes, we assess L3DE's appearance and geometry scores by measuring visual fidelity and spatial accuracy. For dynamic scenes, we focus on validating the motion score by analyzing temporal coherence and movement realism.
Specifically, we use 3D-GS \cite{kerbl20233d} for static scenes and SC-GS \cite{huang2024sc} for dynamic scenes. Rendering quality is quantified using Peak Signal-to-Noise Ratio (PSNR).
To compensate for content-dependent variations in PSNR, we normalize the rendering quality of synthetic videos $Q_{\text{synthetic}}$ relative to that of real videos $Q_{\text{real}}$. This normalization mitigates scene-specific biases, leading to a more robust assessment. The normalized quality difference is defined as :
\begin{equation}
\Delta Q = \max\left(Q_{\text{real}} - Q_{\text{synthetic}}, 0\right).
\end{equation}
To quantify the disparity between real and synthetic videos, we define the simulation gap $G$ based on L3DE scores $S$:
\begin{equation}
G = 1 - S. 
\end{equation}
We then evaluate L3DE’s ability to capture 3D rendering quality by computing the correlation between $\Delta Q$ and $G$. 
As shown in Table \ref{tab:cor_render}, L3DE scores are \textit{positively correlated with 3D rendering quality}, indicating that higher L3DE scores correspond to the superior rendering fidelity. Notably, our L3DE fusion model achieves the highest correlation of $0.7566$, demonstrating strong alignment with the reconstruction-based evaluation.

\vspace{0.05in}\noindent\textbf{L3DE localized region vs. inconsistent region.} 
We assess L3DE’s ability to localize 3D-inconsistent regions in AI-generated videos using the `Generated Videos for In-the-wild Scenes' from the 3D reconstruction verification dataset.
Grad-CAM~\cite{selvaraju2016grad} highlights the regions L3DE focuses on for real-fake classification. To establish reference 3D-inconsistent regions, we split the dataset into training and test sets, ensuring discrepancies are measured only from test viewpoints to mitigate overfitting effects in GS-based reconstruction.
We then quantify the alignment between L3DE-detected regions and rendering-based discrepancy maps. Fig.~\ref{fig:cam_val} presents the quantitative correlation results, demonstrating strong alignment between L3DE-detected and rendering-inconsistent regions. Qualitative comparisons are shown in Fig.~\ref{fig: validation2}.

\subsection{Validation using Human Judgment}
To complement reconstruction-based validation, we conduct human evaluations to assess whether L3DE scores and detected regions align with human perception judgments. This ensures that L3DE not only correlates with objective reconstruction quality but also reflects subjective judgments of 3D visual coherence.

\label{sec:human_align}
\begin{figure}[ht]
    \centering
    \includegraphics[width=0.98\columnwidth]{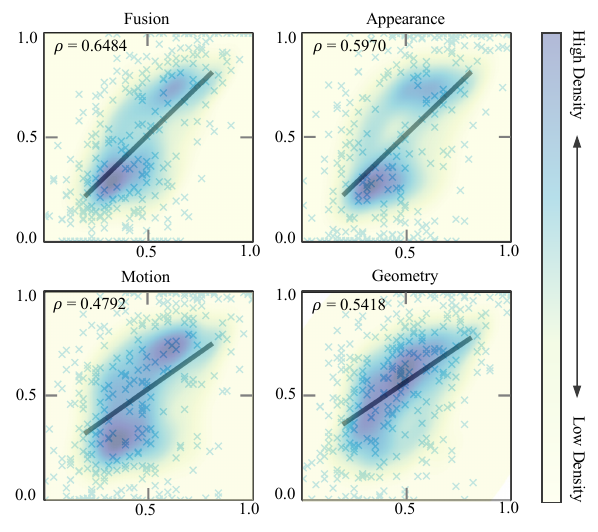}
    \vspace{-5pt}
    \caption{The correlation between L3DE scores and human ratings. The X-axis represents the average human ratings and the Y-axis represents the L3DE scores.}
    \vspace{-8pt}
    \label{fig:cor_human_rating}
\end{figure}

\vspace{0.05in}\noindent\textbf{L3DE scores v.s. human ratings.}
First, We validate the correlation between L3DE scores and human evaluations through a user study involving 15 participants who provided 4,500 annotations on 300 randomly selected AI-generated videos, rating their realism in terms of 3D visual coherence. More details on the study setup are provided in the appendix.  
For each video, we compute the average participant rating as the human rating.  We then evaluate L3DE scores across appearance, geometry, motion, and their fusion. We then compute the Spearman correlation between these scores and the human ratings. 
As shown in Fig.~\ref{fig:cor_human_rating}, L3DE scores exhibit a strong positive correlation with human evaluations, confirming their reliability in assessing generative videos. Notably, the fusion score achieves the highest correlation, underscoring the effectiveness of our fusion strategy.
Additionally, we analyze human ratings for the videos used in the rendering quality experiments and compute their correlation with L3DE scores. It shows that L3DE consistently aligns with both reconstruction quality and human judgment on the same dataset (see Table \ref{tab:cor_render}).

\begin{table}[!h]
    \centering
    \renewcommand\arraystretch{1.1}
    \resizebox{0.8\linewidth}{!}{
    \begin{tabular}{l|ccc}
        \toprule
          & \bf Appearance & \bf Motion & \bf Geometry\\ 
        \midrule
        \bf Average score & 0.8600 & 0.7200 & 0.7400\\
        \midrule
        \bf Spearman's $\rho$ & 0.4894 & 0.4026 & 0.4317\\
        \bottomrule
    \end{tabular}
        }
    \caption{Average human plausibility scores on the Grad-CAM visualization and correlation between L3DE localized region and human-annotated region from different aspects.}
    \vspace{-6pt}
    \label{tab:huamn_ps}
\end{table}

\vspace{0.05in}\noindent\textbf{L3DE localized region v.s. human plausibility.}
We further validate L3DE’s localized regions through an additional user study. 10 volunteers are shown highlighted regions from both L3DE and randomly generated maps, without disclosure of their source to prevent bias. Each participant rates the plausibility of the highlighted regions on a 1–5 scale, with scores subsequently normalized.
As shown in Table~\ref{tab:huamn_ps}, L3DE achieves a significantly higher score (0.7–0.8) compared to random maps (average \textbf{0.21}, minimum 0.2).
To reinforce our validation, we conduct a second experiment where 10 participants annotate unrealistic regions in 30 unseen videos. The correlation between these annotations and L3DE-detected regions (Table~\ref{tab:huamn_ps}) further confirms that L3DE effectively aligns with human perception of unrealistic content.

\begin{table}[t]
    \centering 
    \renewcommand\arraystretch{1.3}
    \resizebox{0.85\linewidth}{!}{
    \begin{tabular}{l|c|c}
        \toprule
        \textbf{Metric} & \textbf{Method} & \textbf{Spearman's $\rho$} \\
        \midrule
        Subject Consistency & VBench \cite{huang2023vbench} & 3.90 \\
        Background Consistency & VBench \cite{huang2023vbench} & 20.68 \\
        Motion Smoothness & VBench \cite{huang2023vbench} & 19.99 \\
        Temporal Consistency & EvalCrafter \cite{liu2024evalcrafter} & 13.85 \\
        L3DE Fusion Score & L3DE & \textbf{64.84} \\
        \bottomrule
    \end{tabular}
    }
    \caption{Correlation of L3DE scores and automatic metrics from different baselines with human ratings.}
    \vspace{-0.4cm}
    \label{tab:comparison}
\end{table}

\section{Analysis and Applications of L3DE}

\subsection{Comparison with Existing Metrics}
While existing methods such as VBench~\cite{huang2023vbench} and EvalCrafter~\cite{liu2024evalcrafter} provide general-purpose video evaluation, they do not specifically assess 3D visual coherence. To compare them with L3DE, we select relevant metrics from each benchmark that focus on spatial and temporal consistency.  
We evaluate them based on their correlation with human judgments, following the standard approach for validating evaluation methods~\cite{huang2023vbench,liu2024evalcrafter}. 
As shown in Table~\ref{tab:comparison}, L3DE achieves a stronger correlation with human ratings than existing metrics, demonstrating its effectiveness in assessing 3D realism in generative videos.  
Beyond correlation analysis, L3DE also introduces unique capabilities, such as identifying unrealistic areas—an aspect missing from existing metrics—which enhances interpretability and provides actionable insights for improving generative models.

\begin{table}[!h]
    \centering 
    \renewcommand\arraystretch{1.3}
    \resizebox{\linewidth}{!}{
    \begin{tabular}{l>{\columncolor{mycolor}}cccc}
        \toprule
        \textbf{Generators} & \textbf{Fusion} & \textbf{Appearance} & \textbf{Motion} & \textbf{Geometry}\\ 
        \midrule
        Runway-Gen3 \cite{runway2024gen3}& 0.7162 & 0.6946 & 0.5768 & 0.6739\\
        MiniMax \cite{minimax2024hailuo}& 0.7932 & 0.7714 & 0.6098 & 0.7251\\
        Vidu \cite{shengshu2024vidu}& 0.7052 & 0.6406 & 0.6228 & \underline{0.7615}\\
        Luma 1.6 \cite{luma2024dm}& 0.5062 & 0.4950 & 0.5853 & 0.6800\\
        Kling 1.5 \cite{kuaishou2024kling} & 0.7518 & 0.7247 & 0.5926 & 0.6927\\
        CogVideoX-5B \cite{yang2024cogvideox} & 0.6104 & 0.5893 & 0.6203 & 0.7539\\
        Sora \cite{videoworldsimulators2024} & \underline{0.8895} & \textbf{0.8394} & \underline{0.6467} & 0.7458\\
        Kling 2.1 \cite{kuaishou2024kling} & \textbf{0.8904} & \underline{0.8129} & \textbf{0.6735} & \textbf{0.7623}\\
        \midrule
        Real Videos & 0.9999 & 0.9950 & 0.8321 & 0.8435\\
        \bottomrule
    \end{tabular}
    }
    \vspace{-0.2cm}
    \caption{Benchmarking results of generative models. The Fusion column, highlighted as the primary L3DE ranking, represents the overall 3D visual coherence. 
    Real videos achieve near-perfect scores, serving as an empirical upper bound for L3DE.}
    \label{tab:evaluation}
\vspace{-0.3cm}
\end{table}

\subsection{Benchmarking Video Generation Models}
Given that L3DE effectively evaluates the 3D visual coherence of generative videos, we expand video generation model benchmarking by introducing 3D visual simulation capabilities as a new assessment dimension, which has been largely overlooked in existing benchmarks. Using the data outlined in Sec.~\ref{sec: data}, we evaluate leading generative models based on their ability to simulate the 3D visual world and present our findings below.

\vspace{0.05in}\noindent\textbf{Quantitative Studies.}
To benchmark generative models, we compute the average L3DE score across all generated videos for each model. The fusion score represents the model’s overall evaluation, while individual scores for appearance, motion, and geometry are also reported.
The evaluation results are shown in Table \ref{tab:evaluation} and the model rankings strongly correlate with large-scale human-preference benchmarks~\cite{VideoArenaLeaderboard} (see appendix), confirming the robustness and generalizability of L3DE.
Based on the overall fusion score, Kling 2.1 \cite{kuaishou2024kling} and Sora \cite{videoworldsimulators2024} produces the highest-quality videos in terms of 3D visual simulation assessment. 
While these models excel in appearance simulation, their motion and geometry scores remain significantly lower, with minimal variation among models.
As a reference, we calculate L3DE scores for a large set of 14,000 real video clips and they achieve an average fusion score of 0.9999, reaffirming the reliability of L3DE. 
Kling's and Sora's fusion and appearance scores exceed 0.8, but their motion and geometry scores are notably lower, indicating potential areas for improvement.
These findings indicate that:
\begin{itemize}
\item While some videos generated by leading models achieve high L3DE scores, most still exhibit significant gaps in 3D visual coherence compared to real videos.
\item The primary distinction among video generation models lies in their ability to simulate appearance, whereas their motion and geometry performance remains notably lower, lacking the fidelity of real-world videos.
\end{itemize}

\vspace{0.05in}\noindent\textbf{Qualitative Studies.}
We analyze the Grad-CAM results from the fusion version of L3DE and observe that, while it provides less direct interpretability compared to individual aspects, it effectively captures more complex artifacts.
For instance, Fusion Grad-CAM effectively identifies physically implausible interactions, such as issues with liquid, glass, and human scaling. For more qualitative studies, please refer to the supplementary.
These findings indicate that integrating multiple cues in L3DE enhances its capability to detect higher-level inconsistencies beyond individual appearance, motion, or geometry assessments.
\subsection{Applications}
We further demonstrate several downstream applications of L3DE, including fake video detection by applying a threshold on the prediction score and enhancing generative video quality by inpainting regions identified by L3DE. More details on these applications can be found in the appendix.

\section{Conclusion and Discussion}
 We present Learned 3D Evaluation (L3DE), a robust and interpretable framework for assessing the 3D visual coherence of generative videos. By leveraging monocular 3D cues—motion, depth, and appearance—from foundation models, L3DE provides an objective and quantifiable measure of discrepancies between real and synthetic videos. Extensive experiments demonstrate L3DE's effectiveness in evaluating videos from generative models, revealing significant 3D simulation gaps and subtle inconsistencies that are often overlooked by human observers. L3DE aligns well with reconstruction quality and human judgment, validating its role as an analytical tool and deepfake detector. Beyond evaluation, L3DE's insights can inform video synthesis improvements, offering a promising avenue for enhancing the realism of AI-generated content. Overall, L3DE presents a powerful tool for advancing our understanding of AI’s capabilities in simulating the 3D visual world, with broad applications in video generation and evaluation.

\vspace{0.2in}\noindent\textbf{Acknowledgments:} 
This work has been supported by Kuaishou Technology, Hong Kong Research Grant Council - Early Career Scheme (Grant No. 27209621), General Research Fund Scheme (Grant No. 17202422), RGC Matching Fund Scheme (RMGS), Lingnan University Start-Up Grant fund code: SUG-001/2526, and Faculty Research Grant fund code:106106. Part of the research work is conducted in the JC STEM Lab of Robotics for Soft Materials funded by The Hong Kong Jockey Club Charities Trust.

{
    \small
    \bibliographystyle{ieeenat_fullname}
    \bibliography{main}
}

\clearpage
\maketitlesupplementary

\renewcommand\thesection{\Alph{section}}
\renewcommand\thefigure{A\arabic{figure}}
\renewcommand\thetable{A\arabic{table}}

\hypersetup{ colorlinks=true,
linkcolor=black,
urlcolor=black,
}



\section{Applications for L3DE}
\label{sec:supp_applications}
In this section, we mainly demonstrate two downstream applications for our proposed L3DE: 1.) \textbf{Fake video detection} and 2.) \textbf{Generative video refinement}.
\subsection{Fake Video Detection}
L3DE is designed to evaluate the 3D real world simulation capabilities of AI-generated videos, enabling it to distinguish low-quality AI-generated videos from real-world ones. Motivated by this capability, we conduct fake video detection experiments to assess how well L3DE performs on this task. This can be achieved by setting a threshold on the L3DE score, allowing us to classify videos as real or fake based on their ability to simulate the real 3D visual world.

Specifically, we use fake videos from our 3D reconstruction verification set together with those from \cite{zeng2024dawn} and an equal number of unseen in-the-wild real videos from Pexels \cite{Pexels} to build a fake video detection benchmark. As there is currently no open-source general fake video detector to the best of our knowledge, we adapt fake image detection methods for videos. To do this, we compare L3DE fusion scores with existing fake image detection methods \cite{wang2023dire,tan2024rethinking,wang2020cnn} by averaging frame-wise predictions to produce a final prediction for each video. The results are presented in Table~\ref{tab:supp_fvd}.

The results indicate that L3DE scores exhibit strong performance in fake video detection, even though L3DE is not specifically designed for this task. Across videos generated by different models, L3DE scores generally achieve higher accuracy than image-based fake detection methods. These results suggest that most synthesized videos still have significant gaps in 3D simulation capabilities. In conclusion, L3DE scores demonstrate strong performance in fake video detection, despite not being specifically designed for this task.
\begin{figure}[t]
    \centering
    \vspace{8pt}
    \includegraphics[width=0.98\columnwidth]{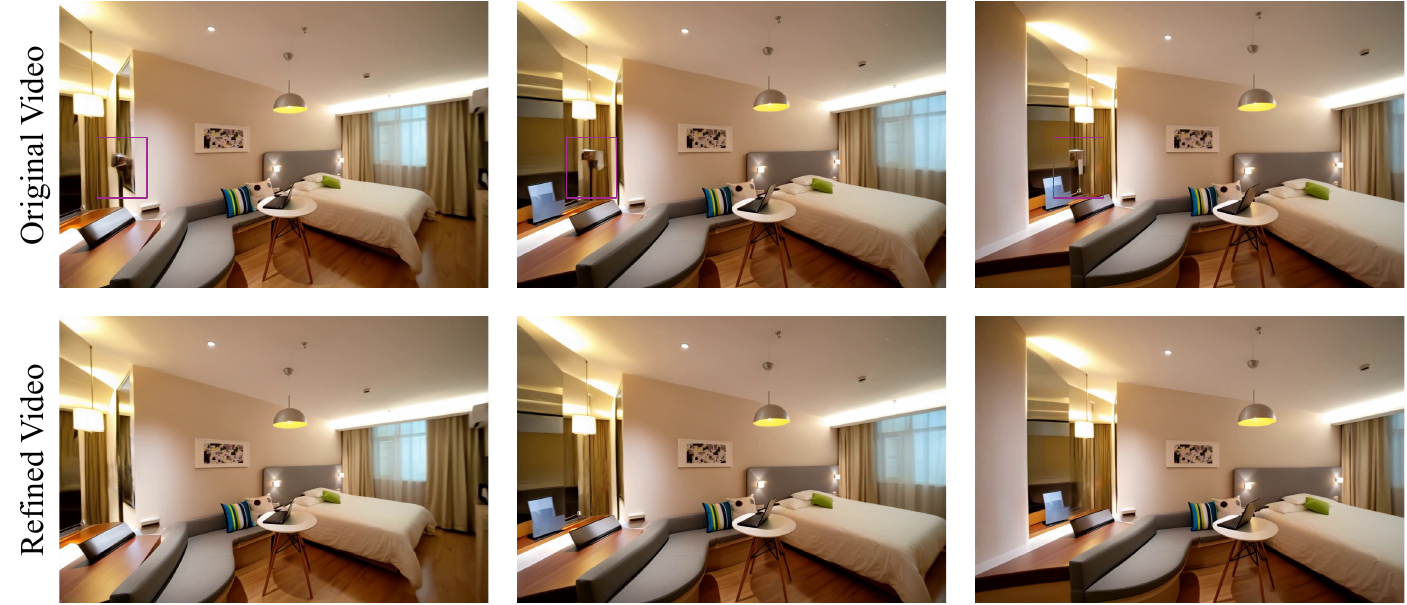}
    \caption{A qualitative result of generative video refinement.  In this example, bounding boxes highlight the regions where artifacts are detected in the original video. After refinement, these artifacts are successfully removed across all frames of the video.}
    \label{fig:supp_refine}
\end{figure}

\begin{figure*}
  \centering
        \includegraphics[width=0.98\linewidth]{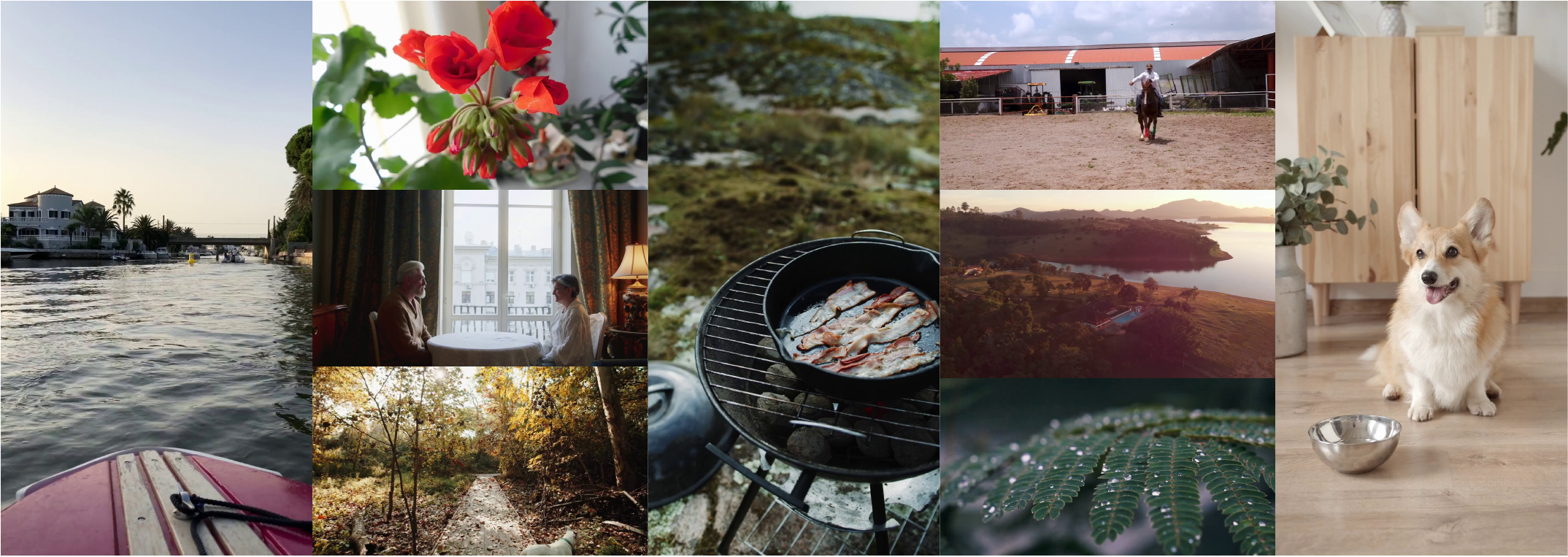}
    \caption{Our collected real-world, in-the-wild videos encompass a wide range of visual content, from indoor to outdoor scenes, including people, animals, landscapes, food, and more.}
    \label{fig:supp_dataset} 
    \vspace{-6pt}
\end{figure*}

\begin{table*}[t]
    \centering
    \resizebox{0.99\linewidth}{!}{%
    \begin{tabular}{l | l | c c c c c c c c}
        \toprule
        Method & Input & $\text{MiniMax}$ & $\text{Kling 1.5}$ & $\text{Runway-Gen3}$ & $\text{Luma Dream Machine}$ & $\text{CogVideoX}$ & $\text{Vidu}$ & $\text{Sora}$ & $\text{Average}$ \\
        \midrule 
        CNNDetection \cite{wang2020cnn} & Image & 49.92 & 50.02 & 50.00 & 50.45 & 50.07 & 50.00 & 49.91 & 50.05  \\
        DIRE \cite{wang2023dire} & Image & 50.00 & 50.00 & 50.00 & 50.00 & 50.00 & 50.00 & 50.00 & 50.00  \\
        NPR \cite{tan2024rethinking} & Image & 60.19 & 67.91 & 64.99 & 54.06 & 35.79 & 36.04 & \textbf{60.82} & 54.25  \\
        \midrule
        L3DE & Video & \textbf{66.51} & \textbf{82.52}  & \textbf{72.19} & \textbf{83.38} & \textbf{76.73} & \textbf{70.01} & 56.31 & \textbf{73.14} \\
        \bottomrule
    \end{tabular}
    }
    \caption{Fake video detection performance of L3DE scores and image-based approaches. The reported metric is accuracy with all values presented as percentages.}
    \label{tab:supp_fvd}
\end{table*}

\subsection{AI-Generated Video Refinement}

In current generative videos with regional artifacts, such artifacts often necessitate discarding the entire video if it does not meet the criteria for downstream tasks. However, with L3DE's ability to identify and localize artifact regions, we can achieve AI-generated video refinement by removing these artifacts in a 3D-consistent manner.

Specifically, we utilize L3DE activation values to localize the regions of artifacts in the keyframes of the downsampled clip. We then employ SAM-2 \cite{ravi2024sam} to refine and propagate the masks across the entire original generative video. Inspired by \cite{mirzaei2023spin}, we implement a 3D-GS-based multi-view consistent inpainting iteratively using LaMa \cite{suvorov2022resolution}.

We demonstrate our results for video refinement in Figure~\ref{fig:supp_refine}. Based on our findings, the artifact-detection capability of L3DE can effectively guide the post-processing step of video refinement, helping to remove artifacts in generative videos.
\section{Data Processing} \label{sec:supp_data}

In this section, we detail our data processing procedures, including raw video processing and video feature extraction.

\subsection{Raw Video Processing}

We follow the approach introduced in \cite{blattmann2023stable} for raw video processing. First, we collect an open-world, in-the-wild long video dataset from Pexels \cite{Pexels}, covering a wide range of content with varying aspect ratios, resolutions, and frame rates. Figure~\ref{fig:supp_dataset} showcases the diversity of our dataset. To avoid biases caused by cuts and fades, we apply PySceneDetect \cite{PySceneDetect} to the long videos.

\begin{figure*}
  \centering
        \includegraphics[width=0.98\linewidth]{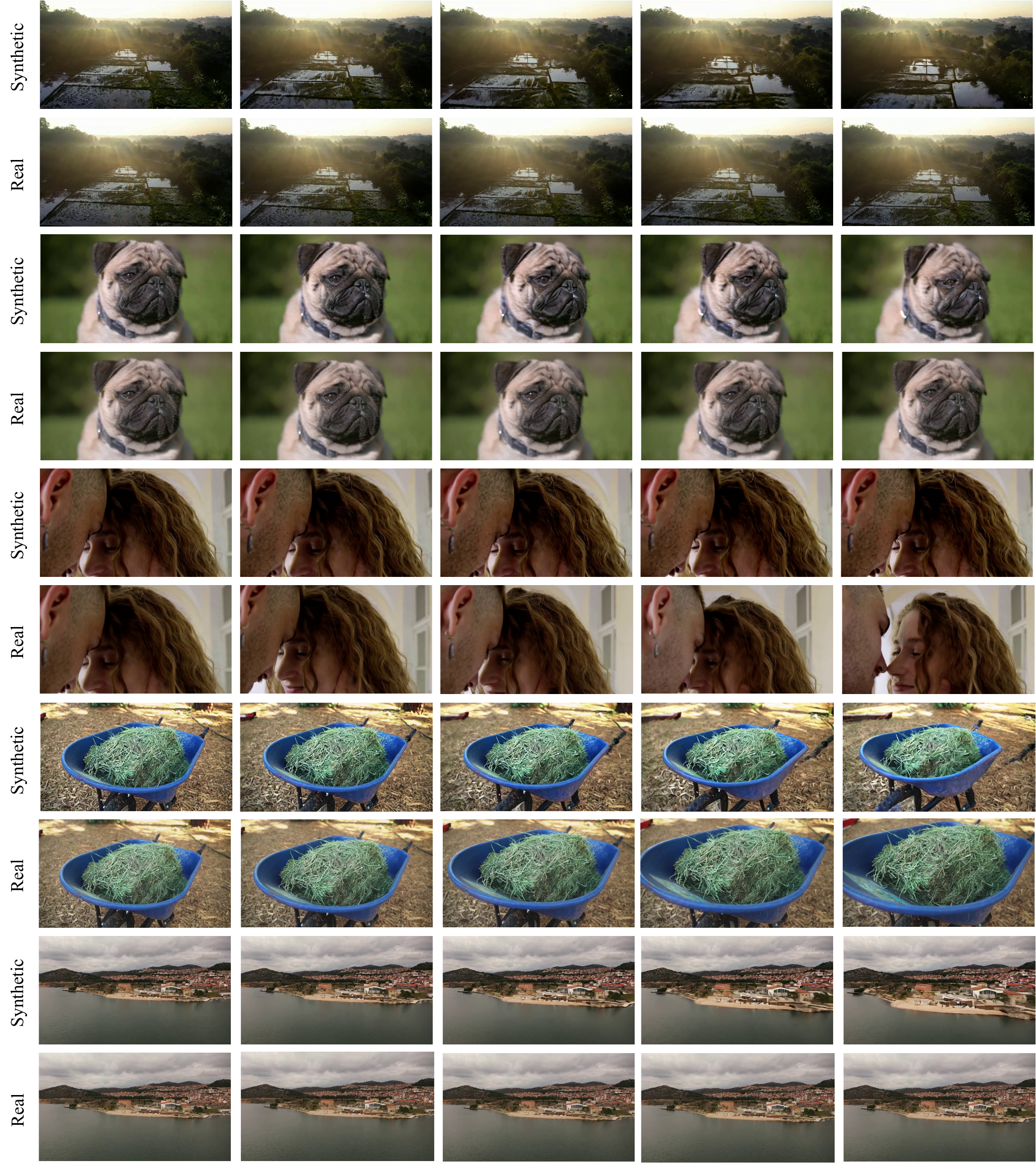}
    \caption{Visualization of randomly sampled paired videos. The images on the left are the image prompts for the generated videos and their first frame. The remaining images show the subsequent frames of the real videos and the generated videos.}
    \label{fig:supp_pair} 
    \vspace{-6pt}
\end{figure*}

Next, to prepare paired data, we slice these videos into equal-length clips of 4 seconds. For videos that do not match the 16:9 aspect ratio, we apply a center crop and resize them to a resolution of $1024 \times 576$ with 25 frames. Additionally, we use the first frame of these processed video clips as image prompts for stable video diffusion \cite{blattmann2023stable} to generate paired synthetic samples. Moreover, we provide visualizations of randomly sampled paired videos in Figure \ref{fig:supp_pair}. As introduced in the main paper, we sample 160,000 paired videos for training the L3DE models.

\subsection{Video Feature Extraction}

We extract video features using different foundation models following their official implementation:
For \textbf{appearance features}, we extract frame-wise features from the DINOv2 ViT-G model \cite{oquab2023dinov2}.
For \textbf{motion features}, we input adjacent frames into RAFT \cite{teed2020raft} to obtain the optical flow sequence of the entire video.
For \textbf{geometry features}, we extract per-frame metric depth using the UniDepth v2 ViT-S model \cite{piccinelli2024unidepth}.

To align the inputs from different proxies, we use the metric depth and DINOv2 features of the first 24 frames of the video clips, since the optical flow maps are calculated based on adjacent frames. This strategy ensures that L3DE simultaneously captures different modalities of 3D proxies.
\begin{figure}[ht]
    \centering
    \includegraphics[width=0.98\columnwidth]{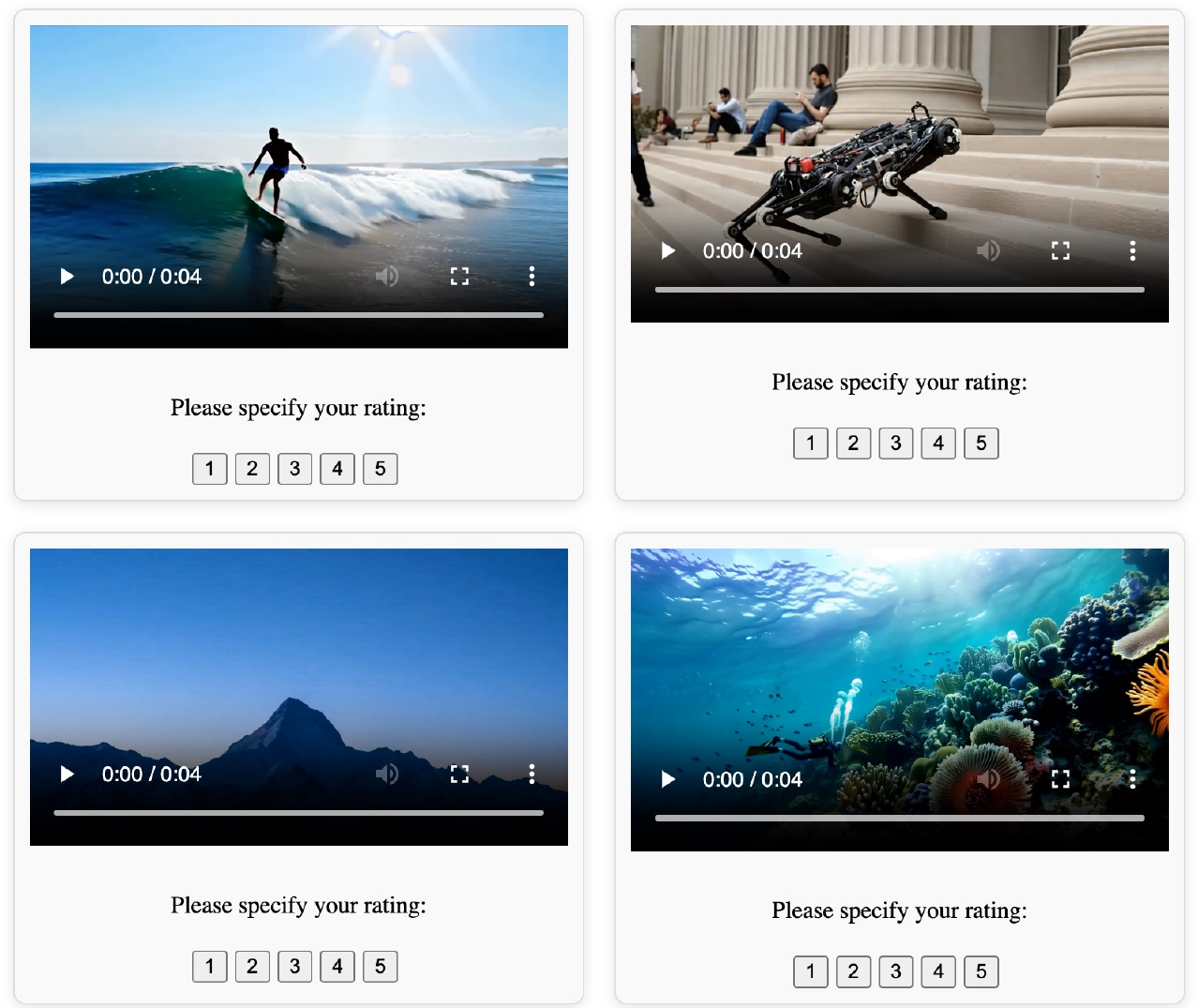}
    \caption{Interface for human rating annotation. Users can provide a rating in the scoring section below after watching the video above.}
    \label{fig:supp_interface}
\end{figure}

\subsection{The Impact of Data Diversity}
Our goal is to construct a diverse training dataset to improve the robustness of L3DE. 
Diversity in training data plays a crucial role in enhancing generalization by exposing the model to a broad range of real-world and challenging scenarios. 
While data diversity can encompass various factors—such as object motion, scene complexity, and environmental variations—we focus on two key aspects in our analysis:  
(1) the role of object motion (\textit{static vs. mixed static-dynamic scenes}) and  
(2) the impact of scene diversity (\textit{indoor-only vs. mixed indoor-outdoor scenes}).  
These controlled experiments illustrate how different types of training data contribute to model performance, reinforcing the importance of a diverse dataset.

\paragraph{Effect of Object Motion}
To assess the impact of object motion, we train two models using different datasets:
one on 10,000 static-scene real and synthetic videos (1:1 ratio), and the other on an equally sized dataset that includes both static and dynamic scenes.
Both models are evaluated on a 2,000-sample test set, which consists of an equal number of real and synthetic videos featuring mixed motion.
The synthetic videos are generated using Kling~\cite{kuaishou2024kling}.
As shown in Table \ref{tab:diversity}, the model trained solely on static scenes underperforms compared to the one trained with motion diversity (\textbf{69.55} \textit{vs.} \textbf{77.70}), confirming that incorporating object motion in training significantly improves generalization.

\paragraph{Effect of Scene Diversity}
To analyze the effect of scene diversity, we train one model using 10,000 indoor real and synthetic videos (1:1 ratio), and another using 10,000 mixed indoor-outdoor videos.
Both models are evaluated on a 2,000-sample mixed indoor-outdoor test set, maintaining a 1:1 ratio of real to synthetic videos.
As seen in Table \ref{tab:diversity}, the model trained only on indoor data exhibits lower accuracy (\textbf{67.60} \textit{vs.} \textbf{76.55}), demonstrating that exposure to a wider variety of environments enhances model robustness.

\begin{table}[t]
    \centering
    \small
    \resizebox{\linewidth}{!}{ 
    \begin{tabular}{lccc}
        \toprule
        \textbf{Experiment} & \textbf{Training Data} & \textbf{Test Data} & \textbf{Accuracy} \\
        \midrule
        \textbf{Static-only} & Static-scene videos & Mixed-motion videos & 69.55 \\
        \textbf{Mixed-motion} & Static + dynamic videos & Mixed-motion videos & 77.70 \\
        \midrule
        \textbf{Indoor-only} & Indoor videos & Mixed indoor-outdoor & 67.60 \\
        \textbf{Indoor + Outdoor} & Indoor + outdoor videos & Mixed indoor-outdoor & 76.55 \\
        \bottomrule
    \end{tabular}
    }
    \caption{Impact of Data Diversity on Model Performance. Training on diverse data significantly improves accuracy.}
    \label{tab:diversity}
\end{table}

\section{L3DE Architecture} \label{sec:supp_arch}

In this section, we provide details about the L3DE architecture, including both the single-proxy and fusion versions.

\subsection{Single-proxy Network}

\begin{figure*}
  \centering
        \includegraphics[width=0.98\linewidth]{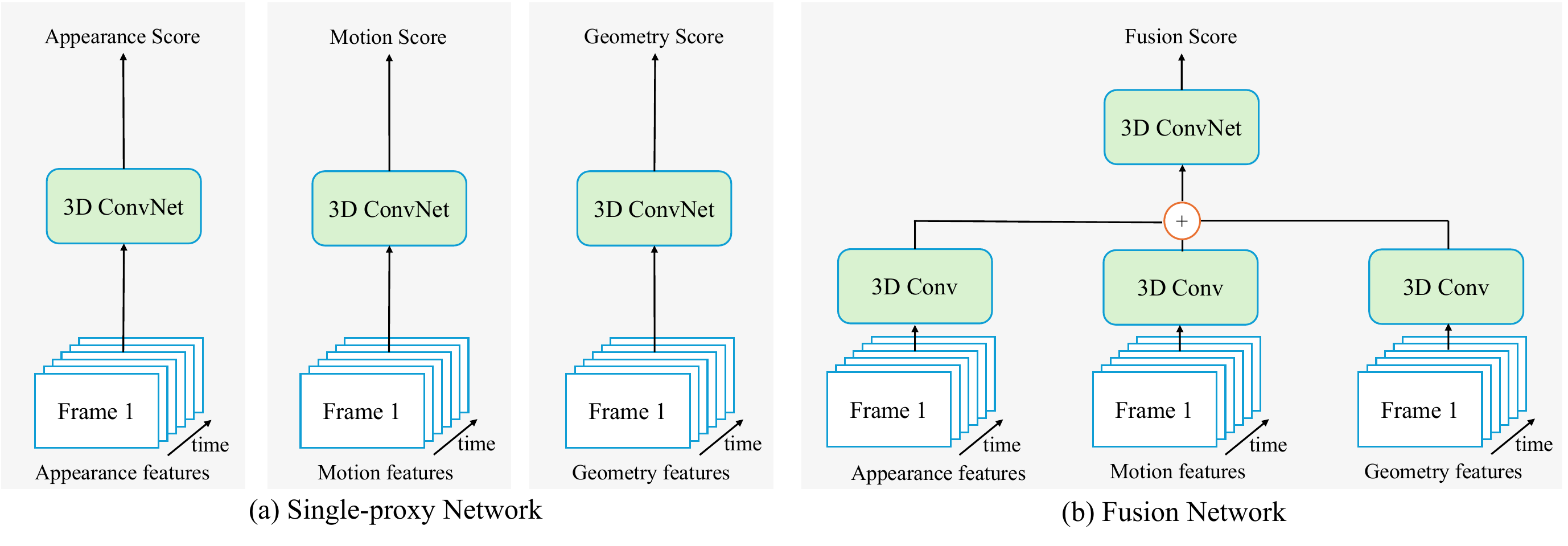}
    \caption{The design of both single-proxy network shown in part (a), and fusion network illustrated in part (b).}
    \label{fig:supp_network} 
    \vspace{-6pt}
\end{figure*}

First, we illustrate our design of the single-proxy version of the L3DE network in Figure~\ref{fig:supp_network}(a). Given a single aspect proxy, such as frame-wise appearance features of a video as input, the 3D ConvNet produces a corresponding confidence score for the video. Specifically, the single-proxy L3DE is a single-branch 3D convolutional network focusing on capturing spatiotemporal features from a single input modality.

The network begins with sequential 3D convolutional layers that progressively encode high-level representations of the input through non-linear activations and feature refinement. After the convolutional stages, the feature map is flattened into a 1D vector, which is passed through a fully connected layer to reduce dimensionality. The final prediction is performed using another fully connected layer with a sigmoid activation, producing a confidence score.

\begin{figure*}
  \centering
        \includegraphics[width=0.98\linewidth]{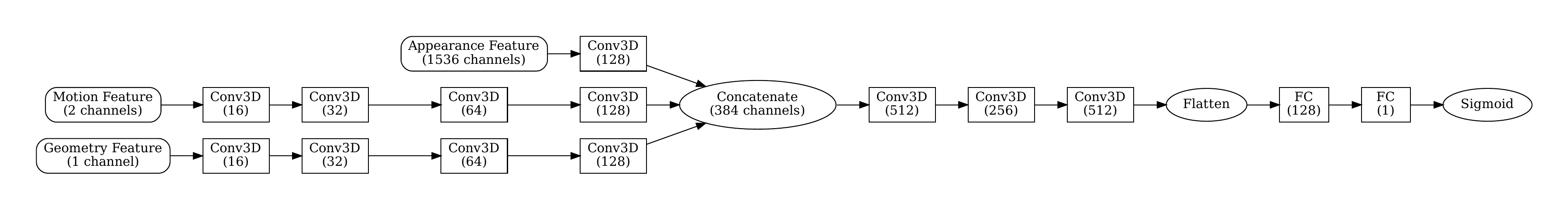}
    \caption{The detailed architecture of fusion network, a 3D convolutional neural network designed for multimodal feature fusion. The network takes three input streams: Appearance Features (1536 channels), Motion Features (2 channels), and Geometry Features (1 channel). Each stream undergoes a series of 3D convolutional layers with ReLU activations before being concatenated into a 384-channel fused representation. The concatenated features are further processed through additional convolutional layers, followed by flattening and fully connected layers.}
    \label{fig:supp_arch} 
    \vspace{-6pt}
\end{figure*}

\subsection{Fusion Network}

Next, we illustrate the design of the fusion version of L3DE in Figure~\ref{fig:supp_network}(b). In detail, the fusion network is a 3D ConvNet integrating appearance, motion, and geometric features through a multi-branch architecture. Each input modality—appearance features, motion features, and geometric features—is processed separately using specialized 3D convolutional layers, which hierarchically encode spatiotemporal information through non-linear activations and down-sampling via strided convolutions.

The outputs of the three branches are concatenated along the channel dimension, enabling the model to jointly leverage complementary features from all modalities, in line with prior efforts~\cite{liu2023mars3d}. The fused representation undergoes further refinement through additional convolutional layers that capture high-level correlations across the integrated features. The network concludes with two fully connected layers and a final sigmoid activation for score prediction.

We also provide the architecture details of the fusion network in Figure~\ref{fig:supp_arch}. Note that each single-branch model adopts the same architecture as its corresponding branch in the fusion network.

\subsection{Ablation Study}
In this section, we conduct an ablation study to analyze the impact of contrastive loss and feature fusion strategies on distinguishing real and synthetic videos in L3DE.
As shown in Table \ref{tab:ablation}, both contrastive loss and fusion strategies play a crucial role in model performance.  
We compare two feature fusion methods:  
(1) Element-wise Addition (Add), where features from different sources are summed component-wise; and  
(2) Feature Concatenation (Concat), where features are stacked along the channel dimension to retain independent information.
First, comparing the Add and Concat fusion strategies, we observe that Concat consistently outperforms Add.  
Without contrastive loss, Concat achieves 68.77\%, surpassing Add (66.01\%), indicating that concatenation preserves richer feature representations.  
When contrastive loss is introduced, performance improves significantly in both fusion strategies (+3.25\% for Add and +4.37\% for Concat), confirming that the loss function enhances feature discrimination.  
Our L3DE setting (Concat + Contrastive Loss) achieves the highest accuracy (73.14\%), as highlighted in Table \ref{tab:ablation}.
These results demonstrate that contrastive loss effectively boosts performance by improving the feature separation between real and synthetic videos.  
Additionally, the superior performance of Concat over Add suggests that maintaining richer feature representations is beneficial for this task.  
Thus, we adopt the Concat + Contrastive Loss setting as the default configuration in L3DE.

\begin{table}[t]
    \renewcommand\arraystretch{1.1}
    \centering
    \small
    \begin{tabular}{c|c|c}
        \toprule
        \bf Fusion Strategy & \bf Contrastive Loss & \bf Accuracy (\%) \\ 
        \midrule
        \multirow{2}{*}{Element-wise Addition} 
        & \xmark & 66.01 \\
        & \cmark & 69.26 \\
        \midrule
        \multirow{2}{*}{Feature Concatenation} 
        & \xmark & 68.77 \\
        & \cellcolor{mygray}\cmark & \cellcolor{mygray}73.14 \\
        \bottomrule
    \end{tabular}
    \caption{Ablation study on contrastive loss and feature fusion strategies (Concat vs. Add). The highlighted row represents our setting and results.}
    \label{tab:ablation}
\end{table}

\subsection{Implementation Details}
We implement our 3D ConvNet using PyTorch \cite{paszke2017automatic}. The models are trained with a learning rate of 1e-4 and a batch size of 20. For video generation with SVD-XT~\cite{blattmann2023stable} and training of L3DE models, we utilize NVIDIA A100 GPUs. Additionally, NVIDIA 4090 GPUs are used for conducting 3D reconstruction experiments.
We follow the official implementation for Grad-CAM~\cite{selvaraju2016grad} visualization.

\section{User Study}
\label{sec:supp_user}




In this section, we provide detailed descriptions of the user studies mentioned in the main paper.

\subsection{User Study for Video Ratings}

We conduct a user study involving 15 volunteers, who provide a total of 4,500 annotations on 300 randomly selected generative videos from our dataset. Annotators are recruited via our internal platform. Participants are aged between 20 and 40, come from diverse educational backgrounds, and do not possess specialized computer vision knowledge, ensuring broad representativeness.

To ensure annotation quality, volunteers complete a pre-labeling task following previous work~\cite{huang2023vbench,liu2024evalcrafter} and only those showing consistent and accurate judgments qualify for the main study.

Qualified participants receive clear scoring guidelines to ensure consistency. The guidelines explicitly instruct them to evaluate the realism of videos based on 3D visual coherence in appearance, motion, and geometry, rather than semantic content or other unrelated factors. Participants rate each video's realism on a 1 to 5 scale, with clear definitions provided:

\begin{itemize}
\item Score \textbf{1}: Videos exhibit obvious visual artifacts, severe geometry deformation, unnatural motion, or evident synthetic features.
\item Score \textbf{2}: Videos have significant artifacts clearly distinguishable from real ones, significantly impacting realism.
\item Score \textbf{3}: Videos contain noticeable but non-disruptive artifacts, moderately realistic overall.
\item Score \textbf{4}: Videos closely resemble real-world footage with minor and infrequent artifacts.
\item Score \textbf{5}: Videos are indistinguishable from real-world footage, exhibiting minimal to no noticeable artifacts or inconsistencies.
\end{itemize}

Participants rate all 300 videos through our internal annotation interface (Figure~\ref{fig:supp_interface}). After collecting annotations, we then compute the Spearman correlation coefficients between these human ratings and the L3DE scores across different modalities. Moreover, to further verify L3DE’s alignment with human perception, we conduct additional human evaluations on the subset "Generated Videos for In-the-wild Scenes." These evaluations comprehensively validate our method’s performance on the same dataset, facilitating comparison with the reconstruction-based validation.

\subsection{User Study for Grad-CAM Region Ratings}




To evaluate the interpretability and effectiveness of the localized regions identified by L3DE (via Grad-CAM), we conduct an additional user study involving 10 qualified volunteers. Participants review 40 randomly selected generative videos from our dataset, each presented alongside visualizations highlighting artifact regions.

Among these 40 videos, for each modality (appearance, motion, and geometry), we randomly select 10 diverse videos. Additionally, we insert 10 videos with randomly generated Grad-CAM highlights serving as a control group to mitigate potential participant biases toward highlighted regions.

Participants view each video along with the corresponding visualization and rate the relevance of highlighted regions to the observed visual artifacts using the following scale:

\begin{itemize}
\item Score \textbf{1}: Highlighted regions are irrelevant or poorly match the perceived artifacts.
\item Score \textbf{2}: Highlighted regions slightly match perceived artifacts but miss major inconsistencies.
\item Score \textbf{3}: Highlighted regions partially match perceived artifacts.
\item Score \textbf{4}: Highlighted regions generally reflect perceived artifacts with minor discrepancies.
\item Score \textbf{5}: Highlighted regions accurately reflect major perceived artifacts.
\end{itemize}

Participants are unaware that 10 of the provided visualizations are randomly highlighted (random baseline) to minimize bias. We specifically evaluate these procedures on the subset "Generated Videos for In-the-wild Scenes" to verify L3DE's effectiveness in localizing artifacts under realistic conditions. Average scores across participants quantify human plausibility, as presented in the main paper. Additionally, 10 participants manually annotate regions they perceive as unrealistic in 30 unseen videos. This serves as a further validation step for Grad-CAM localization, allowing us to quantitatively evaluate pixel-level correlations between human annotations and Grad-CAM highlighted regions.






\section{More Experiments for L3DE}
\label{sec:supp_qualitative}

\subsection{Additional Comparison with Baselines}
To further assess the generalizability of L3DE, we compare its performance against EvalCrafter~\cite{liu2024evalcrafter} using correlation metrics on the EvalCrafter Text-to-Video (ECTV) Dataset.  
EvalCrafter evaluates video quality across multiple dimensions, among which visual quality, motion quality, and temporal consistency are the most relevant to L3DE’s evaluation criteria.
As shown in Table~\ref{tab:ectv}, L3DE achieves a higher correlation with human annotations in terms of visual quality (\textbf{+11.6\%}) and temporal consistency (\textbf{+1.3\%}), demonstrating its strong ability to assess both appearance and temporal coherence.  
L3DE achieves a comparable correlation in motion quality (43.6\% vs. 45.0\%), indicating its effectiveness in capturing motion fidelity. 
These results suggest that L3DE provides a more comprehensive and robust evaluation, particularly in aspects that contribute to overall perceptual quality.

\begin{table}[t]

    \renewcommand\arraystretch{1.1}
    \resizebox{\linewidth}{!}{
    \begin{tabular}{lccc}
        \toprule
        \bf  & \bf Visual Quality & \bf Motion Quality & \bf Temporal Consistency \\ 
        \midrule
        \bf EvalCrafter & 55.4 & \textbf{45.0} & 56.7 \\
        \bf Ours & \textbf{67.0} & 43.6 & \textbf{58.0} \\
        \bottomrule
    \end{tabular}
        }
    \vspace{-0.3cm}
    \caption{Correlation between L3DE scores and human annotations from the ECTV dataset. Appearance, motion, and fusion scores correspond to visual quality, motion quality, and temporal consistency, respectively.}
    \label{tab:ectv}
\vspace{-0.1cm}
\end{table}

\subsection{Comparison with External Human Preference Benchmark}

To further validate the generalizability and robustness of our L3DE results, we compare the ranking of generative video models obtained by L3DE against the publicly available large-scale human preference leaderboard from Video Arena~\cite{VideoArenaLeaderboard}, which aggregates extensive user votes. Although the datasets and specific videos differ, the model rankings obtained by L3DE closely align with those in the Video Arena leaderboard as shown in Table~\ref{tab:supp_arena_comparison}. Notably, both assessments consistently identify similar high-performing and lower-performing generative models. This alignment further confirms that L3DE effectively captures general human perceptual judgments regarding video realism, strengthening the validity of our evaluation framework.

\begin{table}[h]
    \centering
    \resizebox{\linewidth}{!}{
    \begin{tabular}{lccc}
        \toprule
        \textbf{Generative Model} & \textbf{L3DE Score $\uparrow$} & \textbf{Arena ELO $\uparrow$} & \textbf{Ranking (Ours / Arena)} \\
        \midrule
        Sora~\cite{videoworldsimulators2024} & 0.8895 & 1077 & 1 / 1 \\
        MiniMax~\cite{minimax2024hailuo} & 0.7932 & 1067 & 2 / 2 \\
        Kling 1.5~\cite{kuaishou2024kling} & 0.7518 & 1058 & 3 / 3 \\
        Runway-Gen3~\cite{runway2024gen3} & 0.7162 & 1017 & 4 / 4 \\
        CogVideoX~\cite{yang2024cogvideox} & 0.6104 & 811 & 5 / 6 \\
        Luma~\cite{luma2024dm} & 0.5062 & 997 & 6 / 5 \\
        \bottomrule
    \end{tabular}
    }
    \caption{Comparison of generative model rankings obtained by L3DE and human preference judgments from Video Arena~\cite{VideoArenaLeaderboard}. Rankings only consider models appearing in both our 3D visual simulation benchmark and the Video Arena leaderboard. Although datasets differ and there are minor discrepancies in model versions due to rapid iterations in commercial models, the consistent ranking demonstrates L3DE’s alignment with general human perceptual judgments.}
    \label{tab:supp_arena_comparison}
\end{table}

\subsection{More Qualitative Results}
\begin{figure}[t]
    \centering
    \includegraphics[width=0.98\columnwidth]{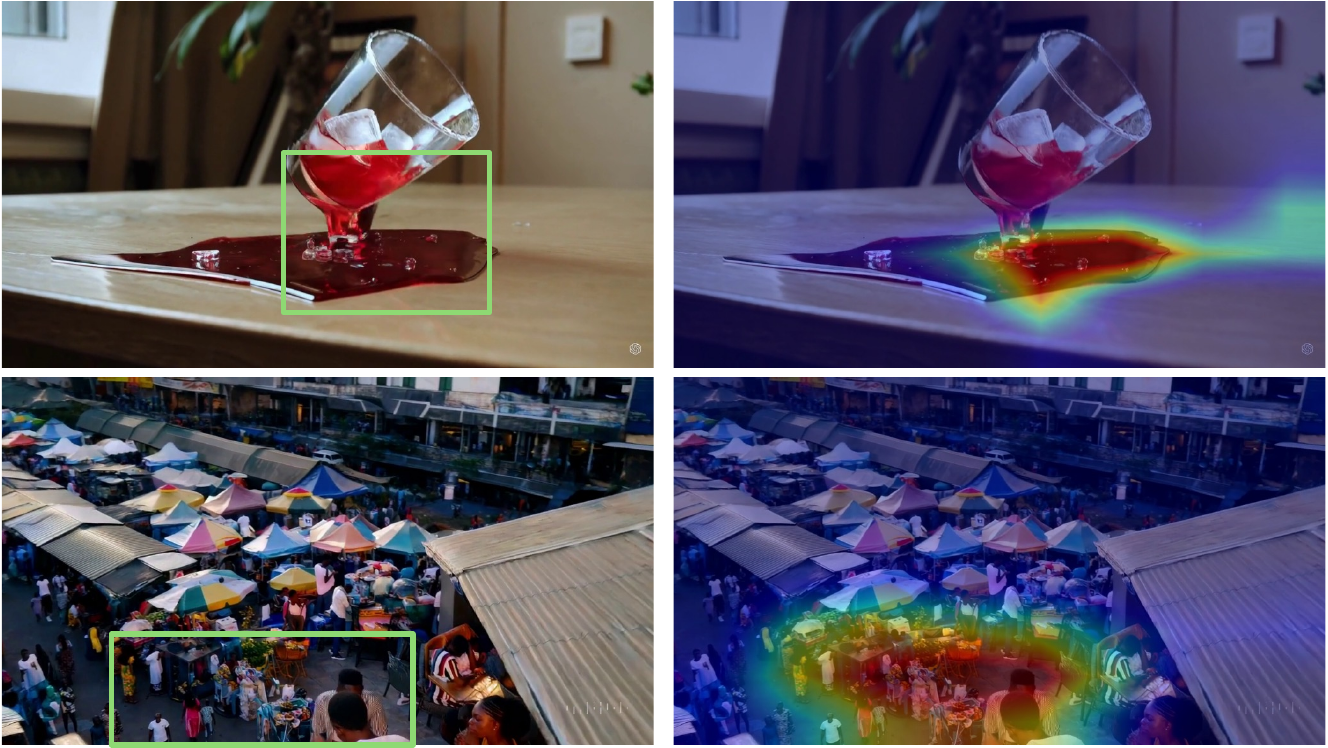}
    \caption{Both clips are from Sora \cite{videoworldsimulators2024}. The first row highlights implausible liquid-glass-table interaction (Score: 0.7256), while the second reveals an incorrect human scale (Score: 0.0023).}
    \vspace{-10pt}
    \label{fig:supp_fusion}
\end{figure}

\begin{figure*}
  \centering
        \includegraphics[width=0.98\linewidth]{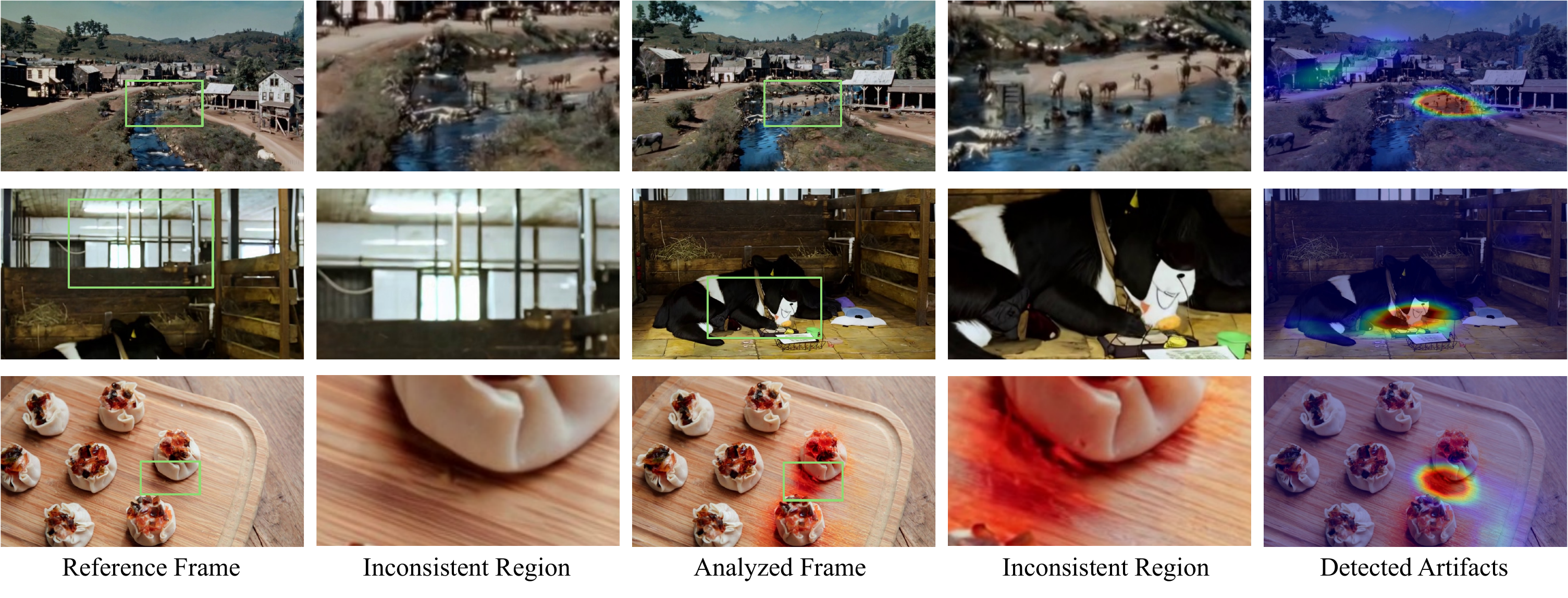}
    \caption{\textbf{Appearance Grad-CAM results of L3DE.} For the first video, appearance Grad-CAM detect unstable scene appearances in the connecting regions between the two scenes, such as objects suddenly appearing or disappearing. For the second video, Appearance Grad-CAM detect regions with inconsistent scene appearance styles. Specifically, the first half of the video depicts a realistic cowshed, but it generates cartoon-style cows inside. For the third video, Appearance Grad-CAM detect a sudden change in the texture of the wooden board and food in the video. More specifically, the color of the wooden board and the food in the marked area change significantly between consecutive frames.}
    \label{fig:supp_appearance} 
    \vspace{-6pt}
\end{figure*}

\begin{figure*}
  \centering
        \includegraphics[width=0.98\linewidth]{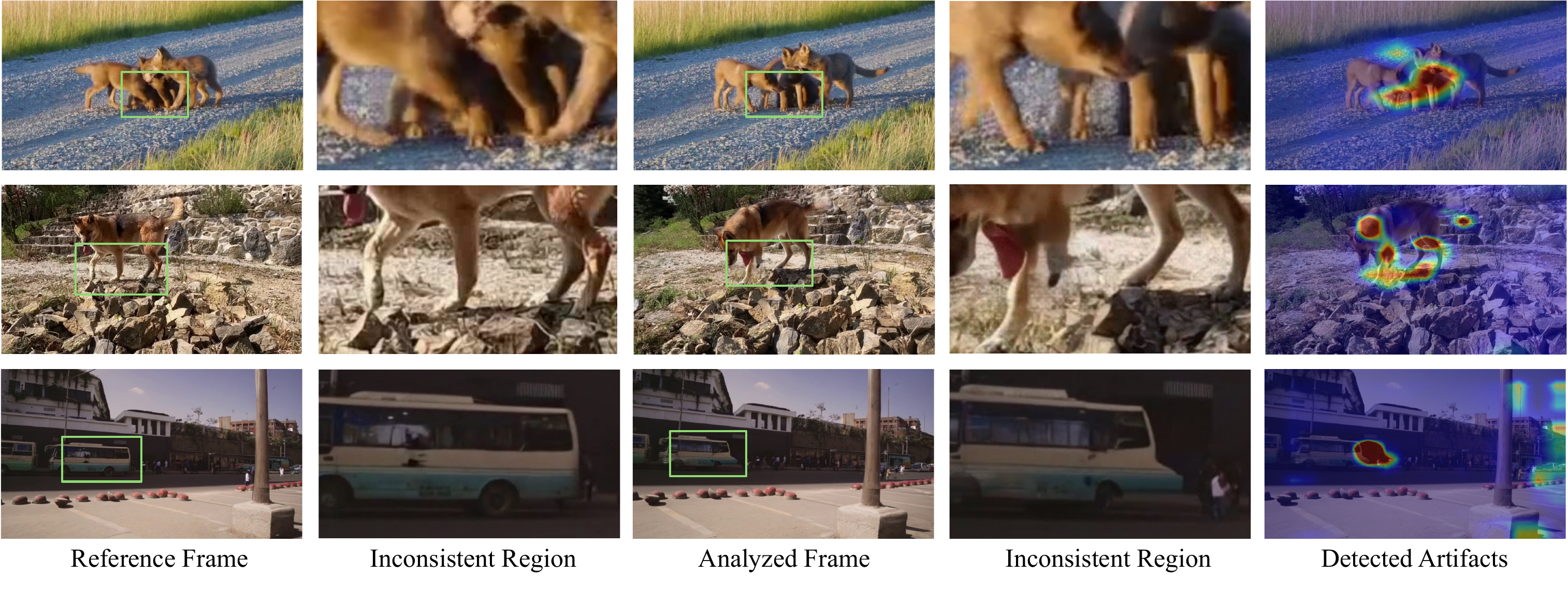}
    \caption{\textbf{Motion Grad-CAM results of L3DE.} For the first video, Motion Grad-CAM detect unnatural motion patterns of the wolves. In the video, the movement of the wolves in the marked area is accompanied by an appearance-disappearance phenomenon, which does not conform to real-world motion patterns. For the second video, Grad-CAM detect regions where the wolf exhibits unnatural motion. Specifically, a wolf that appears with normal four legs in the reference frame experiences sudden disappearance of its legs when moving in subsequent frames. Such motion patterns are inconsistent with real-world ones. For the third video, Grad-CAM detect a sudden unnatural 'compression' motion in the bus, which remain stationary in the first half of the video. This does not conform to real-world motion laws.}
    \label{fig:supp_motion} 
    \vspace{-6pt}
\end{figure*}

\begin{figure*}
  \centering
        \includegraphics[width=0.98\linewidth]{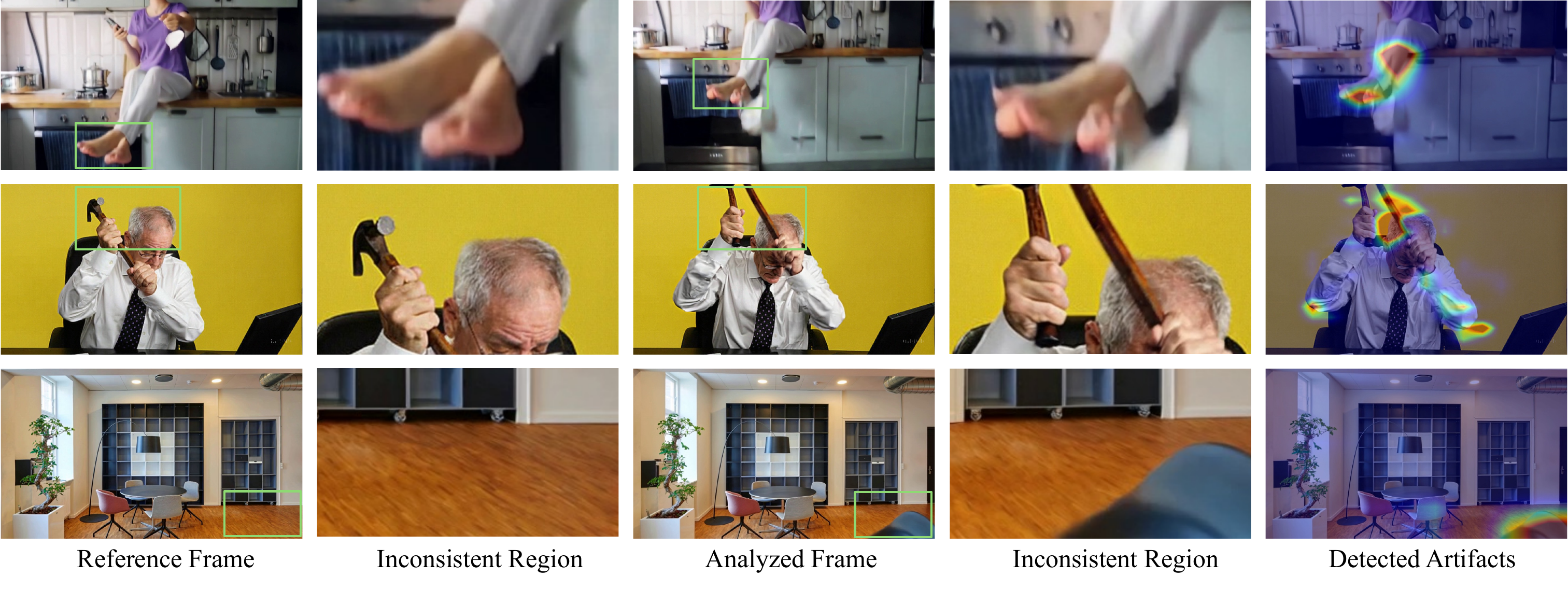}
    \caption{\textbf{Geometry Grad-CAM results of L3DE.} For the first video, Grad-CAM detect inconsistent geometric structures in the person's feet, thereby highlighting the corresponding regions. Specifically, the foot region in the analyzed frame differs from that in the reference frame, exhibiting noticeable blurring and distortion. Such degradation of geometric structure does not conform to real-world patterns. For the second video, Grad-CAM detect an abnormal geometric change in the hammer. In the first half of the video, the elderly person holds a single hammer, but in the subsequent frame, the geometry of the hammer suddenly exhibits a 'cloning' effect, splitting into two. Such geometric inconsistency does not conform to real-world geometry rules. For the third video, Grad-CAM detect regions where a chair suddenly appears in the video. Such sudden changes in scene geometry are inconsistent with real-world patterns.} 
    \label{fig:supp_geometry} 
    \vspace{-6pt}
\end{figure*}

In this section, we provide additional qualitative results of L3DE for reference. Specifically, we illustrate the Grad-CAM results and analyses of L3DE's appearance, motion, and geometry components in Figures~\ref{fig:supp_appearance}, \ref{fig:supp_motion}, and \ref{fig:supp_geometry}, respectively.
We further include comprehensive qualitative examples from the Fusion Grad-CAM analysis, highlighting complex artifacts captured by integrating multiple cues. Figure~\ref{fig:supp_fusion} demonstrates cases involving physically implausible interactions, such as abnormal behaviors of liquids interacting with glass and tables, as well as incorrect human scaling. These examples emphasize the enhanced capability of the fusion model to detect high-level inconsistencies beyond individual appearance, motion, or geometry assessments..

\section{Clarification on Research Scope}
L3DE focuses explicitly on \textbf{3D visual} coherence, specifically assessing appearance, motion, and geometry, as these dimensions are fundamental prerequisites for realistic simulations. It is important to clarify that our method does not comprehensively evaluate all the aspects related to world simulation such as complex interactions (e.g., accurate physics-based interactions, fluid dynamics). Thus, L3DE provides a targeted assessment specifically related to foundational 3D visual coherence, forming a necessary basis for further advancements towards comprehensive world simulation.

\section{Limitations}
\label{sec:supp_limitations}
Although our study takes a very first step to assess the 3D simulation capabilities of AI-generated videos, several challenges remain: 1.) Dataset Size and Diversity: Currently, we use 160000 video clips to train L3DE model. However, the real-world patterns are very complicated and training on more videos will provide a more general and robust evaluation tool. 2.) Limited Generative Video Length: Due to the constraints of current open-source generative video models, which produce relatively short videos, it is challenging to evaluate long-range coherence and object permanence of the future generative videos. To address these limitations, we plan to continually update L3DE to adapt to the generative videos in the future,  and further explore its potential in broader data-centric research~\cite{Zhao_2025_CVPR,tan2023movingonesampleout,tan2024saco, tan2023datapruneinfomax,Zhao_2021_ICCV,liu2024can}.

\end{document}